\documentclass[conference]{IEEEtran}
\IEEEoverridecommandlockouts
\usepackage{cite}
\usepackage{amsmath,amssymb,amsfonts}
\usepackage{etoolbox}
\usepackage{algorithmic}
\usepackage{algorithm}
\usepackage{graphicx}
\usepackage{textcomp}
\usepackage{xcolor}
\usepackage{subfigure}
\usepackage{booktabs}
\usepackage{hyperref}

\def\BibTeX{{\rm B\kern-.05em{\sc i\kern-.025em b}\kern-.08em
    T\kern-.1667em\lower.7ex\hbox{E}\kern-.125emX}}
\begin{document}

\title{	C-DOG: Multi-View Multi-instance Feature Association Using Connected $\delta$-Overlap Graphs
\thanks{Code available at GitHub: https://github.com/Yunghong/C-DOG}
}
\author{Yung-Hong Sun$^{*}$, Ting-Hung Lin, Jiangang Chen, Hongrui Jiang, Yu Hen Hu\\
Department of Electrical and Computer Engineering\\ University of Wisconsin - Madison, WI 53705, USA}

\maketitle

\begin{abstract}
Multi-view multi-instance feature association constitutes a crucial step in 3D reconstruction, facilitating the consistent grouping of object instances across various camera perspectives. The presence of multiple identical objects within a scene often leads to ambiguities for appearance-based feature matching algorithms. Our work circumvents this challenge by exclusively employing geometrical constraints, specifically epipolar geometry, for feature association. We introduce \textbf{C-DOG} (Connected $\delta$-Overlap Graph), an algorithm designed for robust geometrical feature association, even in the presence of noisy feature detections. In a \textbf{C-DOG} graph, two nodes representing 2D feature points from distinct views are connected by an edge if they correspond to the same 3D point. Each edge is weighted by its \textit{epipolar distance}. Ideally, true associations yield a zero distance; however, noisy feature detections can result in non-zero values. To robustly retain edges where the epipolar distance is less than a threshold $\delta$, we employ a Szymkiewicz--Simpson coefficient. This process leads to a $\delta$-neighbor-overlap clustering of 2D nodes. Furthermore, unreliable nodes are pruned from these clusters using an Inter-quartile Range (IQR)-based criterion.
Our extensive experiments on synthetic benchmarks demonstrate that \textbf{C-DOG} not only outperforms geometry-based baseline algorithms but also remains remarkably robust under demanding conditions. This includes scenes with high object density, no visual features, and restricted camera overlap, positioning \textbf{C-DOG} as an excellent solution for scalable 3D reconstruction in practical applications.

\end{abstract}


\section{Introduction}

Multi-view 3D reconstruction in computer vision ~\cite{furukawa2015multi,kubota2007multiview} is the process of generating a three-dimensional model of a scene or object from a collection of two-dimensional images captured from different viewpoints. This capability forms the technical foundation for a wide array of applications, including autonomous driving \cite{teepe2024lifting}, surgery \cite{kunert20133d}, and robotics \cite{aulinas2008slam}, among others. The classical multi-view 3D reconstruction pipeline typically involves several stages: extraction of feature points from 2D images, associating these 2D feature points from different camera views to corresponding 3D points, constructing an approximated 3D surface based on the estimated 3D point cloud, and finally, reconstructing stereo views by rendering the estimated 3D surface with color/texture information observed from the multi-view images.

Among these steps, feature association \cite{huang2024survey} presents the most complex challenge, as it seeks to accurately match hundreds or thousands of 2D points to their corresponding 3D locations. Traditionally, this problem is addressed by first matching the visual appearance surrounding each feature point (using feature descriptors) and then imposing geometric constraints, such as homography \cite{szeliski2022computer}, to simplify computation. It is generally preferable to extract visually distinct features, as this significantly reduces the complexity of feature matching. Recent trends, however, have shifted towards learning-based methods that leverage neural networks to associate objects across views \cite{ji2024view, teepe2024lifting, zhang2025collaborative, shuai2022adaptive, zhang2021direct, zhou2023efficient, ma2022ppt, zhang2021adafuse, yu2022towards}. These approaches often integrate appearance cues---such as texture, color, or shape---with spatial and temporal information to robustly resolve correspondence.

\begin{figure*}
    \centering
    \centerline{\includegraphics[width=1.0\textwidth]{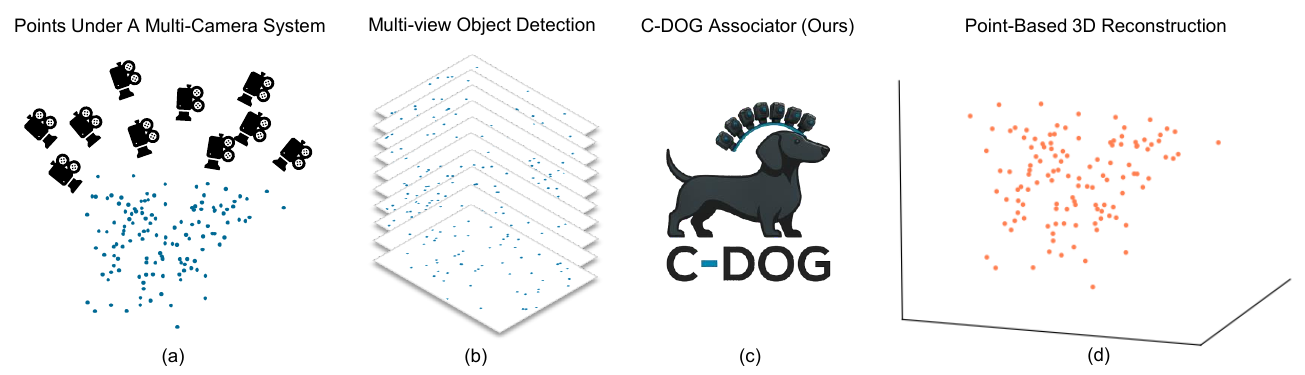}}
    \caption{Demonstration of \textbf{C-DOG} applied to 130 3D points within a multi-camera system from our benchmark. (a) Multiple visually indistinguishable 3D points are observed within a multi-camera setup. (b) Each camera captures 2D projections of these 3D points, which can be obtained using object detectors or keypoint estimators. (c) \textbf{C-DOG} is employed to establish correspondences across views by associating multi-view multi-object observations. (d) The resulting correspondences are then utilized to perform 3D reconstruction.}
    \label{fig:main}
\end{figure*}

\begin{figure}[b]
    \centering
    \centerline{\includegraphics[width=\linewidth]{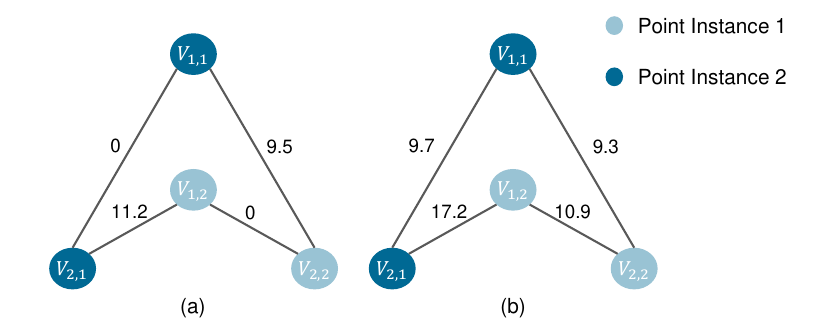}}
    \caption{Example of multi-view multi-object association with two views and two nearby object instances under (a) noise-free and (b) noisy observations. Each node represents a 2D observation in one view. (a) In the ideal case, the association scores between corresponding points (same object across views) are zero, while those between different objects are significantly higher. (b) Under noisy observations, the score distribution becomes ambiguous, making correct correspondence more difficult to establish.}
    \label{fig:err}
\end{figure}

While effective in visually rich and structured environments, existing feature matching approaches often struggle in more challenging scenarios where appearance cues are ambiguous, degraded, or entirely absent. Specifically, scenes containing multiple visually identical instances (multi-instance scenes) pose a fundamental challenge: in such settings, even temporal dynamics may be insufficient to disambiguate object identities across views.

To address these limitations, some approaches abandon appearance features, relying instead solely on multi-view geometric constraints \cite{dong2021shape, cai2020messytable, dong2019fast, chen2014near, fathian2020clear, li2022fast, zhou2015factorized, kahl2025towards, li2015large}. These methods leverage known camera poses---which encode the relative position and orientation of each camera---to infer associations between 2D observations across views. Such geometric relationships are frequently encoded as pairwise affinity scores, indicating the likelihood of correspondence between detections from different views. A common formulation treats each object detection as a graph node, with weighted edges representing the strength of geometric consistency between views.

However, these geometry-only methods are highly sensitive to observation noise and scale poorly with increasing instance counts. As illustrated in Figure~\ref{fig:err}, a noise-free observation in a simple case of two views and two nearby objects yields a distinct association pattern where correct correspondences have zero or near-zero scores. In contrast, under noisy conditions, correct and incorrect associations become nearly indistinguishable. This ambiguity intensifies as the number of objects increases. Although graph-based approaches incorporating epipolar constraints or back-projection errors have been proposed, their performance degrades significantly in the presence of noise and a higher number of objects, thus limiting their reliability in real-world applications.

To overcome these challenges, we propose a novel multi-view, multi-instance feature association method that integrates connected $\delta$-overlap graph (\textbf{C-DOG}) structures with epipolar geometric constraints. This method utilizes a predefined threshold $\delta$ to assess the likelihood of correctness for any association pair, leveraging mutual neighbor connections. Designed to be lightweight and easily integrable, \textbf{C-DOG} functions as a plug-and-play module that bridges object detection or pose estimation with 3D reconstruction, as depicted in Figure~\ref{fig:main}.

Inspired by existing geometry-based multi-view association algorithms, we model each 2D observation from a camera view as an independent node in a graph. Cross-view edges are initially established based on epipolar consistency. We then apply a $\delta$-neighbor-overlap clustering \cite{meghanathan2016greedy} and filtering procedure to construct a graph that captures object correspondences as connected subgraphs, while tolerating missing or noisy edges. This stage filters out erroneous associations based on structural consistency rather than raw epipolar scores, yielding clusters that represent geometrically consistent observations of the same object across views.

To further refine these clusters, we apply an additional filtering step based on the Interquartile Range (IQR) \cite{cetinkayaRundel2024IMS} of edge consistency scores and the 3D back-projection error, which quantifies the triangulation error of matched points. Our approach demonstrates robustness to appearance ambiguity and limited view overlap, and significantly increases the feasible number of objects that can be accurately associated under challenging multi-camera conditions.

Our contributions are summarized as follows:
\begin{itemize}
    \item We propose \textbf{C-DOG}, a novel multi-view multi-object association framework that integrates connected $\delta$-overlap graph modeling with epipolar geometric constraints.
    \item We incorporate an \textbf{IQR-based outlier filtering} mechanism to improve robustness against noisy observations and ambiguous pairwise associations.
    \item Extensive experiments on synthetic benchmarks demonstrate that our method outperforms existing baselines in terms of both precision and completeness.
\end{itemize}

\section{Related Works}
\subsection{Multi-view Feature Matching}
Feature matching serves as a foundational step for higher-level vision tasks, including localization, mapping, and multi-view multi-object association. Classical approaches, such as ORB~\cite{rublee2011orb} and RANSAC~\cite{fischler1981random}, offer computational efficiency and robustness in structured environments but often degrade under wide-baseline viewpoints, severe illumination changes, or texture-poor regions. To address these challenges, deep learning–based matchers such as SuperPoint~\cite{detone2018superpoint}, SuperGlue~\cite{sarlin2020superglue}, and ClusterGNN~\cite{shi2022clustergnn} leverage learned keypoint detectors, descriptors, and context-aware correspondence reasoning. These methods deliver superior performance in challenging scenarios by jointly optimizing detection and matching but incur substantial computational costs and require extensive training data~\cite{huang2024survey}. Recent advances have moved beyond independent pairwise matching: CoMatcher\cite{zhang2025comatcher} introduces GNN-based cross-attention and cross-view geometric constraints to collaboratively reason over multiple co-visible images, improving correspondence reliability in wide-baseline and occluded scenes. Similarly, Roessle and Nießner\cite{roessle2023end2end} integrate differentiable pose optimization with a GNN-based multi-view matcher, enabling pose errors to guide the learning of correspondences and boosting geometric accuracy. In parallel, optimization-driven alternatives such as MATCHEIG\cite{maset2017practical} enforce spectral cycle-consistency using eigen-decomposition to refine noisy pairwise correspondences into globally consistent multi-view matches, while UPMGC-SM\cite{wen2023unpaired} employs structural graph matching and fusion to align unpaired multi-view data without relying on deep features or paired training data. These developments illustrate a spectrum of approaches, from purely geometric and spectral methods to advanced neural architectures, highlighting the trade-off between scalability, accuracy, and computational demand.

\subsection{Deep Learning Multi-view Object Association}
Building upon feature matching, deep learning–based multi-view multi-object association frameworks integrate visual cues, geometric constraints, and temporal information to produce globally consistent associations across views. Many methods adopt Transformer architectures~\cite{vaswani2017attention,dosovitskiy2020image}, whose self-attention mechanisms are effective at capturing interdependencies within and across views~\cite{teepe2024lifting,shuai2022adaptive,zhang2021direct,zhou2023efficient}. To enhance robustness, spatial camera constraints—such as epipolar geometry and homography—are often incorporated to regularize correspondence estimation~\cite{ji2024view,ma2022ppt,zhang2021adafuse}. Beyond static matching, temporal information has been leveraged to maintain consistency across frames: for example, Yu \textit{et al.}\cite{yu2022towards} and Zhang \textit{et al.}\cite{zhang2025collaborative} jointly refine multi-view associations by propagating features from preceding frames. Graph Neural Networks (GNNs)\cite{scarselli2008graph,wu2020comprehensive} are increasingly applied to this problem by modeling detections as graph nodes and formulating correspondence prediction as a structured group prediction task~\cite{wu2021graph,luna2022graph,rodriguez2024multi}. These methods achieve strong performance in visually complex environments by jointly modeling appearance, geometry, and temporal dependencies. However, their success comes with notable limitations: they require large-scale, task-specific labeled datasets for training, are often computationally expensive, and can exhibit reduced robustness in feature-sparse or privacy-sensitive environments. Moreover, when integrated into larger pipelines involving concurrent tasks, their resource demands can exceed available GPU budgets, limiting their deployability in real-time or resource-constrained settings.

\subsection{Multi-view Object Association}
\label{sec:related_work_trainingfree}
To overcome the limitations of deep learning–based approaches, training-free methods provide adaptable alternatives that require no task-specific training data. These approaches typically treat each detected object as a graph node and estimate correspondences using spatial or geometric cues. Greedy strategies~\cite{shafique2005noniterative,lu2004wide} prioritize local edge scores for fast but suboptimal solutions, while global optimization approaches formulate the problem as an energy minimization task, often solved via graph cuts~\cite{kolmogorov2002multi,vogiatzis2005multi}, enabling globally optimal correspondences in settings such as volumetric stereo~\cite{cai2020messytable}. Permutation synchronization~\cite{pachauri2013solving,chen2014near} enforces cross-view consistency by synchronizing partial permutation matrices, with scalable variants such as CLEAR~\cite{fathian2020clear} and its successors~\cite{li2022fast} improving computational efficiency. Factorized Graph Matching (FGM)\cite{zhou2015factorized} reduces the memory and computational overhead of multi-view affinity optimization, while Composition-based Affinity Optimization (CAO)\cite{yan2015multi,swoboda2019convex,kahl2025towards} refines scalability and time complexity for large-scale problems. Extensions like robust bundle adjustment~\cite{zach2014robust} and RKHS-based optimization~\cite{zhang2024rkhs} introduce soft matching and non-linear lifting to better handle outliers and noisy observations. Spectral clustering approaches~\cite{ng2001spectral,li2015large} segment affinity graphs using Laplacian eigenvectors, while connected component clustering~\cite{klasing2008clustering} builds K-NN graphs with adaptive pruning for robust segmentation, as later adapted to 3D human pose estimation~\cite{wu2021graph,cao2017realtime}. Although these approaches may underperform deep learning methods in appearance-rich scenes, they provide lightweight, interpretable, and easily deployable solutions that remain effective in feature-sparse, unstructured, or dynamically changing environments.

\section{Preliminary}
\subsection{Epipolar Geometry}

Let $m \in \{1, 2, \dots, M\}$ be the indices of the camera views, and let $i_m \in \{1, 2, \dots, I(m)\}$ be the indices of the 2D points detected in the $m^\text{th}$ camera view, where $I(m)$ is the indices of 2D points detected in the $m^{th}$ view. 

Let $V_{m,i_m}$ and $V_{m',i_m'}$ be two 2D points on two distinct views $m$ and $m'$, $m \ne m'$ respectively. If these two points are the images {\it associated} with the same 3D point on these two views, their relationship may be described using a {\it fundamental matrix} via the {\it epipolar geometry} \cite{szeliski2022computer}. Specifically, the fundamental matrix between views $m$ and $m'$ is defined as
\begin{equation}
    \mathbf{F}_{m,m'} = \mathbf{K}_m^{-T} [\mathbf{T}_{m,m'}]_\times \mathbf{R}_{m,m'}^{-1} \mathbf{K}_{m'}^{-1}.
\end{equation}
where the intrinsic parameter matrices are denoted by $\mathbf{K}_m$ and $\mathbf{K}_{m'}$, and the extrinsic parameter matrices are denoted by $[\mathbf{R}_m \mid \mathbf{T}_m]$ and $[\mathbf{R}_{m'} \mid \mathbf{T}_{m'}]$ respectively. $[\mathbf{T}_{m,m'}]_\times \in \mathbb{R}^{3 \times 3}$ is a skew-symmetric matrix of the cross-product of translation vector $\mathbf{T}_{m,m'}$. The $3 \times 3$ fundamental matrix has a rank equal to two and is skew-symmetric, namely, 
$$\mathbf{F}_{m,m'} = -\mathbf{F}_{m,m'}^T= -\mathbf{F}_{m',m}$$

\begin{figure}
    \centering
   \centerline{\includegraphics[width=\linewidth]{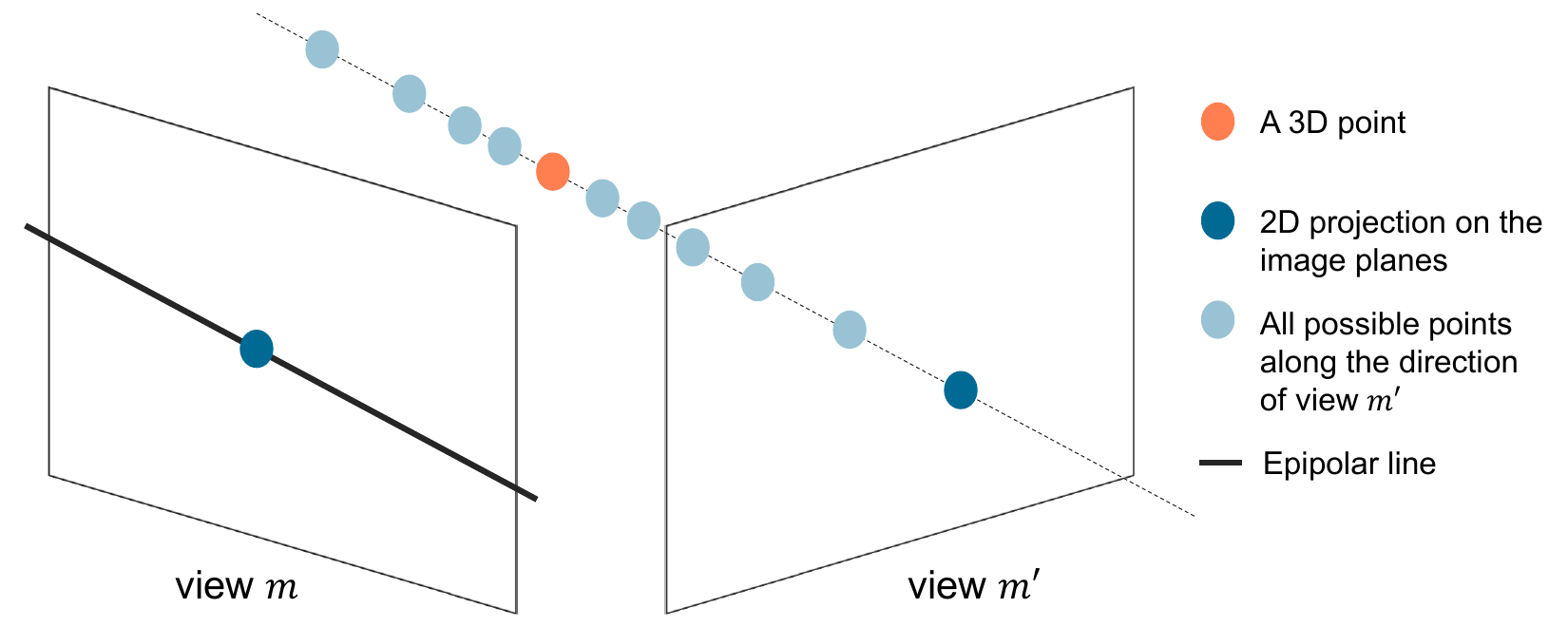}}
    \caption{
    Illustration of the epipolar line concept. Views $m$ and $m'$ observe the same 3D point. All 3D points lying along the viewing ray of $m'$ project to the same image location in view $m'$, forming a 3D line. The projection of this line onto view $m$ yields the epipolar line. In the noise-free case, the corresponding observation in view $m$ should lie exactly on this epipolar line, as both views observe the same underlying 3D point.}
    \label{fig:epipolar}
\end{figure}

Given the fundamental matrix, an {\it epipolar line} on the image plane of the $m'^{th}$ view can be computed: 
\begin{equation}
\label{epiline}
    \mathbf{l}_{m \rightarrow m', i_m} = \mathbf{F}_{m,m'} V_{m,i_m},
\end{equation}
This epipolar line is the image on the $m'^{th}$ view of a ray originated from the camera center of the $m^{th}$ view  to the 3D point corresponding to $V_{m,i_m}$. Since $V_{m',i_{m'}}$ is the image of the same 3D point in the $m'^{th}$ view, it must lies on the epipolar line $\mathbf{l}_{m \rightarrow m', i_m}$. In homogeneous coordinate representation, this implies
\begin{equation}
\label{epidistance}
    V_{m',i_{m'}}^T \mathbf{l}_{m \rightarrow m', i_m} 
    = V_{m,i_{m}}^T \mathbf{l}_{m' \rightarrow m, i_{m'}}
    = 0
\end{equation}
The second expression states that $V_{m,i_m}$ must lie on the epipolar line corresponding to a ray from the camera center of the $m'^{th}$ view to $V_{m',i_{m'}}$ as well as the common 3D point. 
Substituting eq. (\ref{epiline}) into above, one has
\begin{equation}
    V_{m',i_{m'}}^T \mathbf{F}_{m',m} V_{m,i_m} =  
    V_{m,i_m}^T \mathbf{F}_{m,m'} V_{m',i_{m'}}  = 0
\end{equation}
These epipolar geometry constraints are illustrated in 
Fig.~\ref{fig:epipolar}.

{\bf Colinear Case}: If the ray from the camera center of the $m^{th}$ camera to a 3D point also intersect another 3D point. We say the camera center and these two 3D points are colinear. All 3D points that are colinear in the $m^{th}$ view shall cast a single 2D point in that camera. Their images cast on another ($m'^{th}$) view will {\it all} be lying on the same epipolar line associated with that single point in the $m^{th}$ view. Hence, it is possible that for $i'_{m'} \neq i_{m'}$, 
\begin{equation}
\label{colinear}
    V_{m',i'_{m'}}^T \mathbf{F}_{m',m} V_{m,i_m} = V_{m,i_m}^T \mathbf{F}_{m,m'} V_{m',i'_{m'}}  = 0
\end{equation}

{\bf Epipolar Distance}: 
Eq. (\ref{epidistance}) can be interpreted as the Euclidean distance from $V_{m',i_{m'}}$ to the epipolar line $\mathbf{l}_{m \rightarrow m', i_m}$ equals 0 and from $V_{m,i_{m}}$ to the epipolar line $\mathbf{l}_{m' \rightarrow m, i_{m'}}$ equals 0. 
In practice, due to numerical error as well as estimation errors of 2D feature points, the Euclidean distance between a 2D point and corresponding epipolar line may be non-zero. Let us define the {\it epipolar distance} $d_{m,m',i_m,i_{m'}}$ to be the Euclidean distance from a 2D point $V_{m, i_{m}} = [x_o \ y_o \ 1]^T$ (homogeneous coordinate representation) to the epipolar line $\mathbf{l}_{m' \rightarrow m, i_{m'}} = [a \ b \ c]^T$. Then
\begin{equation}
   d_{m,i_m,m',i_{m'}} = \frac{ax_o + by_o + c}{\sqrt{a^2 + b^2}} 
\end{equation}



\subsection{Triangulation}

Given camera intrinsic parameters $\mathbf{K}_m$, and extrinsic parameters $\mathbf{R}_m$, $\mathbf{T}_m$, $m = 1, 2, ...$ and a 3D coordinate $\mathbf{r}$, its projection onto the $m^{th}$ camera plane can be determined by the  pin-hole camera equation \cite{szeliski2022computer}:
\begin{equation}
\label{pinhole}
    V_m = \mathbf{K}_m \left[ \mathbf{R}_m \mid \mathbf{T}_m \right] \left[ \begin{matrix}
    \mathbf{r} \\ 1 \end{matrix} \right] = 
    \mathbf{K}_m \mathbf{R}_m \mathbf{r} + \mathbf{K}_m\mathbf{T}_m 
\end{equation}

Let $\{V_m; m \in G_r \}$ be a set of 2D points where $G_r$ is a subset of indices of camera views such that each 2D point in it is the projection of a 3D point $\mathbf{r}$. Substituting into eq. (\ref{pinhole}), it can be expressed as 
\begin{equation}
\label{triangulation}
    \left[\begin{matrix}
    V_1 \\ \vdots \\ V_{M(\mathbf{r})}
    \end{matrix}\right] = \left[\begin{matrix}
    \mathbf{K}_1 \mathbf{R}_1 \\ \vdots \\ \mathbf{K}_{M(\mathbf{r})} \mathbf{R}_{M(\mathbf{r})}
    \end{matrix}\right] \mathbf{r} +
    \left[\begin{matrix}
    \mathbf{K}_1 \mathbf{T}_1 \\ \vdots \\ \mathbf{K}_{M(\mathbf{r})} \mathbf{T}_{M(\mathbf{r})}
    \end{matrix}\right]
\end{equation}
If there are two or more sets of 2D observations, then the 3D coordinate $\mathbf{r}$ may be estimated from eq. (\ref{triangulation}) by minimizing the least square estimation error.

\subsection{Back-Projection Error}

Denote the estimated 3D coordinate as $\hat{\mathbf{r}}$. One may then {\it back-project} this coordinate to the image planes using the pin-hole camera equation eq. (\ref{pinhole}) to yield a set of back-projected 2D coordinates $\{\hat{V}_m; m \in M(\mathbf{r})\}$. 

Ideally, one would expect
$V_m = \hat{V}_m; m \in M(\mathbf{r})$. However, in practice, their difference is often non-zero. One may define a {\it back projection error} (BPE) for a specific 3D point $\mathbf{r}$ as:
\begin{equation}
\label{BPEr}
    BPE(\mathbf{r}) = \frac{1}{|G_r|} \sum_{m \in G_r} \|V_{m,i_m} - \hat{V}_{m,i_m}\|^2
\end{equation}

BPE may be various due to a number of potential causes: estimation error of 2D feature points, calibration error of camera poses (intrinsic and extrinsic parameters), and feature association error. Nonetheless, it is a metric that rewards consistency between observation (2D feature points) and the model prediction (estimated 3D coordinates view calibrated camera models). Thus, a performance metric to evaluate the quality of multiview feature matching would be the averaged back-projection error 
\begin{equation}
\label{BPE}
    BPE = \frac{1}{L} \sum_{\mathbf{r}} BPE(\mathbf{r})
\end{equation}
where $L$ is the total number of detected 3D points whose coordinates are estimated from associated 2D point using triangulation.

\section{Methodology}
\begin{figure*}[ht]
\centerline{\includegraphics[width=1.0\textwidth]{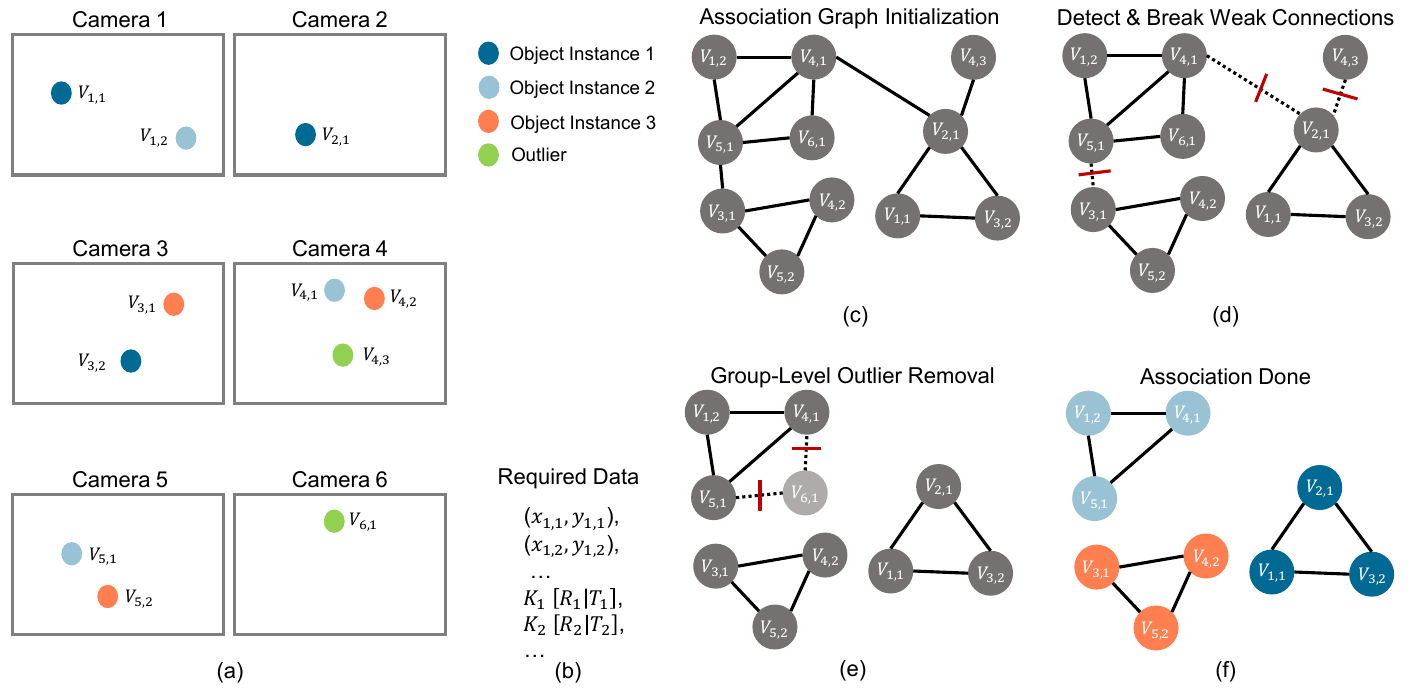}}
\caption{Overview of C-DOG with an illustrative example with three 3D points and 6 cameras:  
(a) A multi-camera system observes object points from multiple viewpoints. Outliers may exist in the observation. 
(b) Required inputs include the camera pose for each view and 2D observed points in pixel coordinates.  
(c) Initial graph generation: each node represents a 2D point in a specific view, and edges between nodes are weighted by epipolar distances. Only view-wise minimal connections that satisfy the epipolar constraint are retained.  
(d) Weak connection removal: sparse inter-group connections are identified and eliminated based on node neighborhood overlap density ~\cite{meghanathan2016greedy}, as they typically represent incorrect associations.  
(e) Group-level outlier removal: outliers are detected using 3D back-projection error within each group.  
(f) Final output: associated groups of 2D points across views, where each group corresponds to a distinct 3D object instance.}
\label{epipolarMatching}
\end{figure*}

\subsection{Problem Formulation}

In this work, we assume $M$ calibrated cameras (views) are used to capture images of unknown number of 3D points. A 3D point needs not appear in the field of views of every cameras. A feature detection (and localization) algorithm will is used to detect and estimate the 2D coordinate of the image of 3D points. The goal is to correctly estimate the number and 3D coordinates of 3D points. The averaged square back projection error is used as an indirect performance metric of the accuracy of the estimated 3D coordinate. 

To estimate the number and 3D coordinates of the 3D points, one must associate the 2D feature points detected at multiple views to corresponding 3D point before the triangulation step described in eq. (\ref{triangulation}) is applied. Unlike existing feature association methods \cite{ji2024view,teepe2024lifting,zhang2025collaborative,shuai2022adaptive,zhang2021direct,zhou2023efficient,ma2022ppt,zhang2021adafuse,yu2022towards}, here no feature descriptor is available. Hence, the feature association must be solely based on epipolar geometry. This leads to the following 2D Point feature matching problem:

{\bf Point Feature Matching Problem} 
Give a calibrated multi-camera system so that the intrinsic and extrinsic camera poses are known.
Let $\{V_{m,i_m}; 1 \le m \le M, 1 \le i_m \le I(m) \}$ be a set of 2D feature points where $m$ is the index of camera views, $i_m$ is the point index within the $m^{th}$ view and $I(m)$ is the number of 2D points within the $m^{th}$ view. We assume these 2D feature points are images of $K$ 3D points. 
The objective of the point feature matching problem is to group these 2D feature points into $L$ {\it association groups} $\{G_{\ell}; 1 \le \ell \le L\}$, $L \le K$ so that the 2D feature points within each association group are associated with the same 3D point. This grouping must satisfy the following constraints:
\begin{enumerate}
    \item $2 \le |G_{\ell}| \le M$ where $|G_{\ell}|$ is the number of members (size) in the set $G_{\ell}$.
    \item If $V_{m,i_m} \in G_{\ell}$ and $V_{m',i_{m'}} \in G_{\ell}$, then $m \neq m'$.  
\end{enumerate}
The quality of the grouping shall be evaluated using the averaged square back projection error defined in eq. (\ref{BPEr}) and eq. (\ref{BPE}). 

In this work, we will solve this point feature matching problem without assuming any feature descriptor is available. If one evaluate the epipolar distance between a 2D point $V_{m,i_m}$ in the $m^{th}$ view and {\it every} 2D feature points $V_{m',i_{m'}}$ in the $m'^{th}$ camera view, one may expect
\begin{equation}
    d_{m,i_m,m',i_{m'}} \ge 0
\end{equation}
If for a specific $i^*_{m'}$ such that 
\begin{equation}
\label{matched}
    d_{m,i_m,m',i^*_{m'}} = 0
\end{equation}
then one may conclude that $V_{m,i_m}$ and $V_{m',i^*_{m'}}$ are a pair of feature points associated with the same 3D point. Hence both of them should be assigned to the same association group. By repeating this process for every 2D points in every camera view, a desired grouping for the Point Feature Matching Problem may be obtained. From eq. (\ref{colinear}) we note that when two or more 3D points are {\it colinear} with respect to the camera center of the $m^{th}$ view, then we would also have
\begin{equation}
\label{matched2}
    d_{m,i_m,m',i'^*_{m'}} = 0
\end{equation}

We define {\it outliers} as 2D feature points that are not included in any association groups. An outlier may correspond to a 3D point which appears in only one camera view or an extremely noisy 2D feature point. An outlier 2D feature point will not be used for triangulation. 

The association groups may be represented graphically: Denote each 2D feature point as a vertex in a graph, and assign an edge between a pair of vertices if both are associated to the same 3D point. As such, all the vertices in an association group form a {\it clique} because there will be an edge between any pair of vertices of this group. 

Ideally procedures discussed so far should be able to solve this Point Feature Matching Problem. However, this is under the assumption that the 2D feature points observations are made without any modeling (camera calibration model) error nor any numerical estimation error during triangulation and back projection. In other words, in the ideal situation, $E_{BP} = 0$. In practice, noisy observations and camera models are expected. hence, a robust feature association algorithm that is less sensitive to the observation and modeling noise is desired. Below, we propose a robust algorithm that leverages a Connected $\delta$-Overlap Graph (C-DOG) representation and epipolar geometry to solve the point feature matching problem.

\subsection{C-DOG Overview}


\textbf{C-DOG} is a geometry-aware association framework designed to serve as an intermediate module between 2D object detection (or pose estimation) and 3D reconstruction. 
In \textbf{C-DOG}, each 2D detection in a camera view is represented as a node in a graph. An edge will be established between a pair of nodes if these nodes are likely associated with the same 3D point. 
As illustrated in Fig.~\ref{epipolarMatching}, the C-DOG consists of four main stages:
\begin{enumerate}
    \item \textbf{Association graph initialization}: construct candidate connections using epipolar geometry.
    \item \textbf{Weak edge pruning}: detect and eliminate unreliable links based on $\delta$-overlap constraints.
    \item \textbf{Group-level outlier removal}: refine groups using Interquartile Range (IQR) filtering and 3D back-projection error.
\end{enumerate}
\subsection{Initial Association and Connection Graph Generation}

\begin{figure}
    \centering
   \centerline{\includegraphics[width=\linewidth]{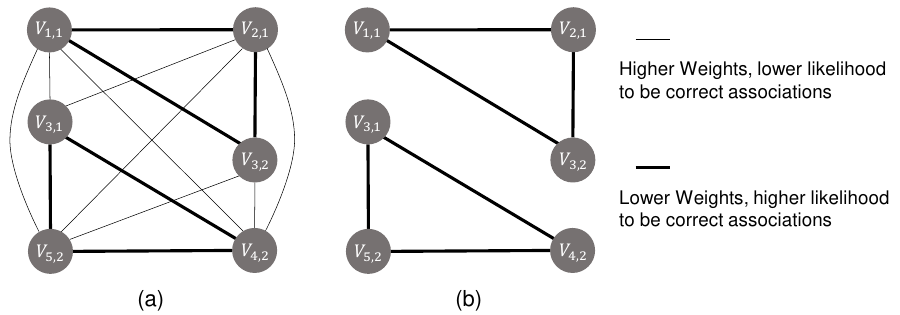}}
    \caption{An example of association graph initialization using object instances 2 and 3 from Fig.~\ref{epipolarMatching}. (a) Epipolar geometry is applied to each pair of points to compute edge weights. Node pairs with lower edge weights (epipolar scores) are more likely to be correct associations. (b) Edges with higher weights are removed, forming several groups. This completes the initialization of the association graph.}
    \label{fig:ini}
\end{figure}

To construct the initial association graph, we leverage epipolar geometry to assess geometric consistency across camera views. By calculating the epipolar distance for each point pair as shown in Section III(a), we initialize a connectivity graph where each 2D point in each view is represented as a node that belongs to an instance, and edges between nodes are determined by their corresponding epipolar distances. As illustrated in Fig.~\ref{fig:ini}, each point is evaluated against all points from other views, and edges with large epipolar distances are subsequently pruned. Algorithm~\ref{alg:graph_initialization} outlines the procedure for constructing the initial graph. Each point $X_{m,i}$ is represented as node $V_{m,i}$ in the graph system.

\begin{algorithm}
\caption{Association Graph Initialization}
\label{alg:graph_initialization}
\begin{algorithmic}[1]
\REQUIRE  Camera intrinsics $\{K_m\}_{m=1}^M$, rotations $\{R_m\}_{m=1}^M$, and translations $\{T_m\}_{m=1}^M$; 2D point (nodes) $\mathcal{V} = \left\{ V_{m,i_m} \;\middle|\; m = 1, \dots, M;\ i_m = 1, \dots, I(m) \right\}$  in pixel coordinates; Epipolar distance threshold $\tau$
\ENSURE Initial graph $\mathcal{G} = (\mathcal{V}, \mathcal{E})$

\FOR{each view $m \in \{1, \dots, M\}$}
    \FOR{each view $m' \in \{1, \dots, M\}$ where $m' \neq m$}
        \FOR{each node $V_{m,i_m}$ in view $m$}
            \STATE Compute the epipolar line $\mathbf{l}_{m \rightarrow m', i_m}$ 
            \FOR{each node $V_{m',i_{m'}}$ in view $m'$}
                \STATE Compute epipolar distance $d_{m,m',i_m,i_{m'}}$
            \ENDFOR
            \STATE Find $i_{m'}^* = \arg\min_{i_{m'}} d_{m,m',i_m,i_{m'}}$
            \IF{$d_{m,m',i_m,i_{m'}^*} < \tau$}
                \STATE Add directional edge $\mathcal{E}_{m,i_m \rightarrow m',i_{m'}^*} = V_{m,i_m} \rightarrow V_{m',i_{m'}^*}$ to $\mathcal{E}$ with weights $d_{m,m',i_m,i_{m'}^*}$

            \ENDIF
        \ENDFOR
    \ENDFOR
\ENDFOR
\STATE Summarize connected component $\mathcal{G}_k \subseteq \mathcal{G}$

\RETURN $\mathcal{G}$
\end{algorithmic}
\end{algorithm}

For each point $X_{m,i_m}$ in view $m$, we compute its epipolar distance to all points $X_{m',i_{m'}}$ in every other view $m' \neq m$. For each such view $m'$, we retain the point with the minimum distance that satisfies a geometric consistency threshold $\tau$:
\begin{equation}
    i_{m'}^* = \arg\min_{i_{m'}} d_{m,m',i_m,i_{m'}} \quad \text{s.t.} \quad d_{m,m',i_m,i_{m'}^*} < \tau.
\end{equation}
If no point satisfies this condition, no correspondence is added for that view.

Thresholding provides an effective mechanism for discarding clearly incorrect point correspondences. The threshold $\tau$ can be estimated based on the level of observation noise, assuming the 2D points are perturbed by zero-mean Gaussian noise. That is, each observed point $\hat{X}_{m,i_m}$ is modeled as a noisy version of the true point: 
\begin{equation}
    \hat{X}_{m,i_m} = X_{m,i_m} + \epsilon, \quad \epsilon \sim \mathcal{N}(0, \sigma^2),
\end{equation}
where $\epsilon$ denotes zero-mean Gaussian noise with standard deviation $\sigma$.

Since the epipolar distance involves comparisons between two independently noisy observations, the combined noise follows a Gaussian distribution with standard deviation $\sqrt{2}\sigma$. In practical situations, the noise standard deviation $\sigma$ can be reliably estimated from 2D observations using statistical methods ~\cite{cetinkayaRundel2024IMS}, assuming the noise is zero-mean and follows a Gaussian distribution. By applying the empirical rule, we set the threshold using a scale factor $\alpha$, where typical choices like $\alpha = 2$ capture approximately 95\% of the noise variation in one dimension.

Thus, the threshold is defined as:
\begin{equation}
    \tau = \alpha \sqrt{2} \sigma,
\end{equation}
where $\alpha$ controls the tolerance to noise.

After this step, all point pairs have been traversed, resulting in an initial bidirectional connection graph composed of multiple connected groups. Each connected component represents a potential correspondence, indicating that the 2D points from different views within the group likely correspond to the same underlying 3D object instance.

\subsection{Weak Edge Pruning}


Once the association graph is initialized, the next step is to identify and eliminate erroneously connected point pairs. Specifically, we evaluate the structural consistency of each edge based on the neighborhood overlap between the connected nodes. Rather than relying on edge weights (e.g., epipolar distances) and edge directions, we utilize graph connectivity information to distinguish between strong and weak connections. Although epipolar geometry serves as a principled foundation for initializing associations, its effectiveness diminishes in the presence of noise and viewpoint variation, limiting its reliability for downstream decision-making. Strong connections are indicative of likely correct associations, whereas weak connections suggest potential mismatches. By removing weak edges, we aim to prune erroneous links and improve the purity of the graph's internal grouping.

In a correctly associated group, the connections are typically dense: each node is expected to be directly connected to most or all other nodes within the group. In contrast, inter-group connections—introduced by noise or ambiguity—are sparse. This structural distinction motivates a neighborhood-overlap-based pruning strategy: when a wrong connection bridges two distinct object groups, it typically results in low mutual connectivity. Removing such weak connections helps isolate distinct object instances.

To quantify the connection strength between two nodes \( V_{m,i} \) and \( V_{m',i'} \), we introduce a neighborhood-overlap score inspired by the Szymkiewicz–Simpson coefficient~\cite{hennig2015handbook}:
\begin{equation}
    \theta(V_{m,i}, V_{m',i_{m'}}) = \frac{\left| N[V_{m,i}] \cap N[V_{m',i_{m'}}] \right|}{\max\left(\left| N[V_{m,i}] \right|, \left| N[V_{m',i_{m'}}] \right| \right)},
\end{equation}
where $N[V_{m,i}]$ denotes the closed neighborhood of \( V_{m,i} \)—i.e., the node and its directly connected neighbors. A high score indicates that both nodes share a large number of common neighbors, suggesting consistency in their local graph structures. Including the node itself in the neighborhood computation helps mitigate score underestimation in cases where the number of nodes is limited. For example, if only two nodes are connected and present in a group, the open neighborhood intersection yields a score of zero, while the closed neighborhood yields a score of one, appropriately reflecting the fact that the two nodes form a densely connected group of size two. This adjustment similarly benefits small groups with only a few nodes, where sparse connectivity can otherwise lead to misleadingly low overlap scores.

Instead of the minimum used in the Szymkiewicz–Simpson coefficient, the maximum is incorporated in the denominator to ensure a conservative estimate: mismatches with imbalanced neighborhood sizes receive lower scores, introducing a higher penalty and improving robustness to incorrect associations. Such that, $0<\theta(.)\leq 1$. In the ideal case, where both nodes are correctly associated within the same group, the score approaches 1. Given a threshold $\delta$, a connection is classified as strong if $\theta(V_{m,i}, V_{m',i'}) > \delta$, and as weak if $\theta(V_{m,i}, V_{m',i'}) \leq \delta$. Through grid-search on our benchmark, we found that setting $\delta \in [0.5,0.55]$ yields the best overall performance. This threshold is intuitive, as it requires that at least half of each node's neighborhood overlaps with the other, ensuring that a majority consensus supports the connection before it is classified as strong.

\begin{algorithm}[t!]
\caption{Weak Edge Pruning}
\label{alg:weak_pruning}
\begin{algorithmic}[1]
\REQUIRE Initial graph \( \mathcal{G} = (\mathcal{V}, \mathcal{E}) \)
\ENSURE Filtered graph \( \mathcal{G}_{pruned} = (\mathcal{V}, \mathcal{E}_{\text{strong}}) \)

\FOR{each connected component \( \mathcal{G}_k \subseteq \mathcal{G} \)}
    \FOR{each connected pair \( (V_{k,m,i}, V_{k,m',i'}) \in \mathcal{G}_k \)}
        \STATE Compute overlap score \( \theta(V_{k,m,i}, V_{k,m',i'}) \)
        \IF{ \( \theta(V_{k,m,i}, V_{k,m',i'}) < \delta \) }
            \STATE Remove weak edge \( (V_{k,m,i}, V_{k,m',i'}) \) from \( \mathcal{E} \)
        \ENDIF
    \ENDFOR
\ENDFOR
\STATE Summarize the connected componenst as groups
\RETURN Pruned graph \( \mathcal{G}_{pruned} \)
\end{algorithmic}
\end{algorithm}

Through the Algorithm \ref{alg:weak_pruning}, by retaining only strong connections, we preserve coherent groups in which nodes are densely connected and structurally consistent. This step significantly enhances the quality of the association graph by separating likely object instances and eliminating ambiguous or spurious links.

\subsection{Group-level Outlier Removal}
While pruning weak connections effectively separates loosely connected or mismatched components, certain outliers may still remain within groups. These outliers may satisfy the epipolar constraint and retain structurally strong connections, making them difficult to detect using connectivity-based measures alone. To further refine group consistency, we introduce a statistically driven outlier removal strategy that leverages 3D reconstruction and back-projection consistency across views as shown in Section III(b).

Specifically, for each group containing three or more nodes, we consider all pairs of 2D observations to reconstruct candidate 3D points. Each reconstructed point is then back-projected onto the remaining views in the same group. The discrepancy between a back-projected point and its corresponding observed point in each view is measured as the back-projection error (BPE). Ideally, if all observations are correctly associated, the BPE should remain low. By aggregating these errors, we compute a per-node average BPE, which serves as a quantitative score for identifying outliers.

\begin{algorithm}[t!]
\caption{Group-level Outlier Removal}
\label{alg:outlier_removal}
\begin{algorithmic}[1]
\REQUIRE Refined graph $\mathcal{G}_{pruned} = (\mathcal{V}, \mathcal{E}_{strong})$
\ENSURE Graph $\mathcal{G}_{rm}=(\mathcal{V}, \mathcal{E}_{rm})$ with outlier nodes' edges removed

\FOR{each group $G_k \subseteq \mathcal{G}_p$}
    \REPEAT
        \FOR{each pair of nodes $V_{m,i}$ and $V_{m',i_{m'}}$ in $\mathcal{V}_k$}
            \STATE Triangulate 3D point $X^{3D}_l$ from $X_{m,i}$ and $X_{m',i_{m'}}$
            \FOR{each remaining node $V_{m'',i_{m''}} \subseteq \mathcal{V}_k$, where $m'' \ne m, m'$}
                \STATE Back-project $X^{3D}_l$ to view $m''$ as $\hat{X}_{m'',i_{m''}}$
                \STATE Compute back-projection error $\mathrm{BPE}_{m,m',m''}$
            \ENDFOR
        \ENDFOR
        \STATE Compute average BPE for each node $V \subseteq \mathcal{V}_k$
        \STATE Apply IQR-based filtering to identify outliers
        \STATE Remove outlier nodes from $\mathcal{V}_k$ and corresponding edges from $\mathcal{E}_{strong}$
    \UNTIL{no outliers are detected in $G_k$}
\ENDFOR

\RETURN Graph $\mathcal{G}_{rm}$ with cleaned groups
\end{algorithmic}
\end{algorithm}

For each group, we select all valid node pairs, reconstruct their corresponding 3D points, and compute the BPE in each remaining view. We then average the BPE values for each node across all reconstructions and back-projections, resulting in a single mean BPE score per node. 

To identify outliers, we apply an IQR filter to the distribution of average BPE scores. Let $Q_1$ and $Q_3$ denote the 25th and 75th percentiles, respectively. The IQR is defined as:
\begin{equation}
    \mathrm{IQR} = Q_3 - Q_1.
\end{equation}

The acceptable BPE range is then bounded by:
\begin{equation}
    \begin{aligned}
    \mathrm{lb} &= Q_1 - \alpha \cdot \mathrm{IQR} \\
    \mathrm{ub} &= Q_3 + \alpha \cdot \mathrm{IQR},
    \end{aligned}
\end{equation}

where $\alpha$ is a hyperparameter that controls the strictness of outlier rejection (set to $\alpha = 2$ in our experiments). Given that the minimum possible BPE is zero, which indicates a perfect projection, we incorporate this prior by padding the list of average BPE values with a zero before computing percentiles and explicitly enforcing the lower bound as zero. This adjustment increases sensitivity in scenarios where the BPE values are generally high, allowing the method to better distinguish relatively large errors. A node is considered an outlier and removed from the group if its average BPE falls outside the interval $[\mathrm{lb}, \mathrm{ub}]$. This step helps minimize BPE, thereby improving point association consistency and enhancing the accuracy of 3D reconstruction.


Algorithm \ref{alg:outlier_removal} outlines the detailed procedure, where each group is iteratively examined for outliers until none are detected. This refinement step yields cleaner groups with fewer outlier-associated nodes, thereby reducing reconstruction errors and enhancing the overall consistency of the multi-view association.

\subsection{Error Association Group Removal}
While the previous step effectively removes intra-group outliers, some erroneous associations may still remain. Because the noise level of the algorithm is unknown, the acceptable BPE range cannot be predefined. To address this, we evaluate each group’s average BPE score, as defined in Equation~\ref{BPE}, and remove entire groups when necessary to avoid incorporating erroneous reconstructions. Specifically, we first compute the BPE for each group and sort these scores in ascending order. We then examine consecutive groups, and if the difference between two successive BPE values exhibits a sudden increase, we discard the remaining groups. This approach provides an adaptive mechanism for identifying and removing erroneous association groups under unknown noise conditions.

\subsection{Synthetic Benchmark Generation}

To evaluate the proposed method in a controlled and featureless setting, we design a synthetic benchmark that eliminates confounding factors such as texture, keypoint structures, and semantic cues. This benchmark isolates the core challenges of multi-view multi-object association by minimizing all variables except for camera poses and 2D observation coordinates. Our goal is to assess the method's performance under idealized conditions, with a focus solely on geometric consistency.

Unlike popular datasets \cite{belagiannis20143d,joo2015panoptic}, which include rich semantic and structural cues, our benchmark omits any explicit feature information or high-level spatial priors, such as object articulated poses. Instead, it consists solely of 3D points and their 2D projections across multiple views. We employ ten calibrated camera views obtained from the EasyVis system~\cite{sun2024easyvis2,sun2025easyvis}, where camera parameters are estimated using SfM~\cite{moulon2017openmvg}. Each camera is represented by its intrinsic matrix $K$, rotation matrix $R$, and translation vector $T$. These matrices describe the orientation and position of a camera in 3D space, and the rule of projection of 3D points to the 2D camera view in pixel coordinates.

The dataset consists of 3D point instances, ranging from 1 to 130, with increments of 1 from 1 to 20 and 5 from 20 to 130. Each number of point instance is generated in 5 batches, and each batch is unique. Each instance is generated from a uniform 3D distribution. The corresponding 2D projections are computed for all ten calibrated views, after which Gaussian noise is added to the 2D points independently in the $x$ and $y$ axes. The noise standard deviation varies from 0 to 5 pixels, in increments of 0.25. In total, the benchmark contains 9,575 unique 3D point groups and 46,192 associated 2D projections for each noise level, spanning a comprehensive range of noise levels and scene complexities. The dataset does not simulate missing detections or false positives and each 3D point has a corresponding 2D observation in all views.

This fully synthetic setup allows precise control over the number of 3D points, the level of observation noise, and the spatial distribution of points. Moreover, it avoids the need for time-consuming manual annotations, which become increasingly impractical as the number of points and views increases. Automatic access to ground-truth associations is particularly critical in our setting for calculating evaluation scores.

\section{Experiments}
\begin{table*}
\centering
\label{ablation study}
\caption{Ablation study of performance under varying numbers of camera views with Gaussian noise $\sigma = 3.00$.}
\begin{tabular}{lcccccccccccc}
\toprule
\textbf{\# views} & \textbf{$\mathrm{G}$-$F_1$} & \textbf{ $\mathrm{G}$-$\mathrm{IoU}$} & \textbf{ $\mathrm{mP}$-$P$} & \textbf{ $\mathrm{mP}$-$R$} & \textbf{$\mathrm{mP}$-$F_1$} & \textbf{ $\mathrm{mP}$-$\mathrm{IoU}$} & \textbf{$\mathrm{PG}$-$P$} & \textbf{$\mathrm{PG}$-$R$}& \textbf{$\mathrm{PG}$-$F_1$}  &\textbf{$3DErr$} & \textbf{$bpe$} & {time(ms)}\\
\midrule
2 views & 0.836 & 0.789 & 0.782 & 0.782 & 0.782 & 0.782 & 0.819 & 0.726 & 0.766 & 0.050 & 9.651 & 1.704\\
3 views & 0.861 & 0.792 & 0.770 & 0.746 & 0.754 & 0.738 & 0.794 & 0.676 & 0.726 & 0.153 & 20.157 & 14.835\\
4 views & 0.743 & 0.639 & 0.612 & 0.598 & 0.601 & 0.581 & 0.679 & 0.501 & 0.562 & 0.343 & 36.743 & 26.760\\
5 views & 0.775 & 0.674 & 0.637 & 0.612 & 0.618 & 0.591 & 0.718 & 0.540 & 0.601 & 0.368 & 38.126 & 75.601\\
6 views & 0.855 & 0.767 & 0.721 & 0.664 & 0.684 & 0.648 & 0.753 & 0.612 & 0.668 & 0.480 & 44.305 & 195.533 \\
7 views & 0.875 & 0.790 & 0.732 & 0.662 & 0.686 & 0.646 & 0.765 & 0.628 & 0.685 & 0.550 & 29.571 & 324.526 \\
8 views & 0.890 & 0.811 & 0.753 & 0.680 & 0.706 & 0.666 & 0.797 & 0.662 & 0.719 & 0.455 & 25.259 & 471.898\\
9 views & 0.880 & 0.793 & 0.746 & 0.710 & 0.721 & 0.699 & 0.838 & 0.681 & 0.747 & 0.828 & 21.033 & 498.642\\
10 views & 0.881 & 0.796 & 0.751 & 0.715 & 0.727 & 0.706 & 0.854 & 0.694 & 0.761 & 0.702 & 15.585 & 523.373\\
\bottomrule
\end{tabular}
\end{table*}
\subsection{Experimental Setup}
We constructed a synthetic benchmark by generating random 3D points uniformly distributed in space and projecting them onto 10 camera views using known camera poses. Gaussian noise was added to the 2D projections with standard deviations ranging from 0 to 5 pixels, in increments of 0.25. The number of 3D points varies from 1 to 130: from 1 to 20 in steps of 1, and from 20 to 130 in steps of 5. The camera parameters were estimated from a real multi-camera system~\cite{sun2024easyvis2} using SfM~\cite{moulon2017openmvg}. To evaluate robustness under varying view availability, we randomly drop a subset of views, retaining between 2 and 9 views in increments of 1. For each subset, excluding the full 10-view case, the selected views are randomly sampled from the original set of 10 calibrated views.

All experiments were conducted on a desktop equipped with an NVIDIA RTX 3080 Ti GPU, an Intel i9-12900K CPU, and 32 GB of RAM. The implementation was developed in Python.

The evaluation metrics are based on true positives (TP), false positives (FP), and false negatives (FN), whose definitions vary depending on whether the evaluation is performed at the group level or the point level.

We adopt the following standard evaluation metrics: precision ($P = \mathrm{TP}/(\mathrm{TP} + \mathrm{FP})$), recall ($R = \mathrm{TP}/(\mathrm{TP} + \mathrm{FN})$), $F_1$ score ($F_1 = 2PR / (P + R)$), and intersection over union ($\mathrm{IoU} = \mathrm{TP}/(\mathrm{TP} + \mathrm{FP} + \mathrm{FN})$).

Based on these formulations, we report three kinds of scores with different definitions of TP, FP, and FN:

\textbf{Group-level Scores ($\mathrm{G}$-):} All predicted groups are evaluated, with isolated nodes treated as separate groups. $\mathrm{TP}$ is the number of predicted groups that match ground-truth groups. $\mathrm{FP}$ is the number of predicted groups that do not match any ground-truth group. $\mathrm{FN}$ is the number of ground-truth groups that are not matched by any prediction. This score reflects the system’s overall ability to associate groups correctly, allowing for reasonable matching slack.

\textbf{Mean Point Scores ($\mathrm{mP}$-):} For each predicted group with valid observations (i.e., from at least two views), we compute point-level precision. A point is counted as a $\mathrm{TP}$ if it is correctly associated with the dominant ground-truth group label, a $\mathrm{FP}$ if wrongly associated, and a $\mathrm{FN}$ if missed. The group class is determined by the majority of its associated ground-truth points. We compute precision per group and report the average across all such groups. This score reflects the method’s fine-grained association performance—i.e., how accurately each group aggregates its points.

\textbf{Perfect Group Scores ($\mathrm{PG}$-):} A predicted group is considered a $\mathrm{TP}$ if it is perfectly associated—that is, it contains only the points from one ground-truth instance. Imperfect or duplicated groups are counted as $\mathrm{FP}$, and missed ground-truth groups are counted as $\mathrm{FN}$. This score reflects the method’s ability to produce entirely correct groupings. In practical applications, having a larger number of perfectly associated groups is crucial for achieving high-quality and interpretable 3D reconstructions.

We denote final scores by combining the prefix and metric symbol.We report: $\mathrm{G}$-$F_1$, $\mathrm{G}$-$\mathrm{IoU}$, $\mathrm{mP}$-$P$, $\mathrm{mP}$-$R$, $\mathrm{mP}$-$F_1$, $\mathrm{mP}$-$\mathrm{IoU}$, $\mathrm{PG}$-$P$, $\mathrm{PG}$-$R$, and $\mathrm{PG}$-$F_1$. A higher score indicates better performance.

In addition, to evaluate the quality of 3D reconstruction, we compute the 3D reconstruction error ($\mathrm{3DErr}$) and back-projection error ($\mathrm{BPE}$). $\mathrm{3DErr}$ is defined as the average Euclidean distance between the reconstructed 3D points and their corresponding ground-truth positions, where reconstruction is performed using points within each associated group. $\mathrm{BPE}$ is calculated by projecting each reconstructed 3D point onto all valid ground-truth 2D views, and measuring the Euclidean distance between the back-projected points and the corresponding ground-truth 2D observations in pixel space. Lower values of $\mathrm{3DErr}$ and $\mathrm{BPE}$ indicate higher association accuracy and reconstruction quality.

Finally, we assess runtime performance by measuring the average time required to complete one full association process. This provides insight into the computational cost of each evaluated method.

\subsection{Ablation Study}

\begin{table*}
\centering
\caption{Quantitative comparison of all methods under Gaussian noise with standard deviation $\sigma = 0.00$.}
\label{tab:compare_noise000}
\begin{tabular}{lcccccccccccc}
\toprule
\textbf{Method} & \textbf{$\mathrm{G}$-$F_1$} & \textbf{ $\mathrm{G}$-$\mathrm{IoU}$} & \textbf{ $\mathrm{mP}$-$P$} & \textbf{ $\mathrm{mP}$-$R$} & \textbf{$\mathrm{mP}$-$F_1$} & \textbf{ $\mathrm{mP}$-$\mathrm{IoU}$} & \textbf{$\mathrm{PG}$-$P$} & \textbf{$\mathrm{PG}$-$R$}& \textbf{$\mathrm{PG}$-$F_1$}  &\textbf{$3DErr$} & \textbf{$bpe$} & {time(ms)}\\
\midrule
Greedy & 0.218 & 0.161 & 0.138 &  0.162& 0.145 & 0.138 & 0.326 & 0.111 & 0.137 & 0.464& 87.516 & 102.733\\
CAO & 0.606 & 0.496 & 0.383 & 0.164 & 0.229 & 0.163 & 0.566 & 0.367 & 0.383 & 0.056 & 6.227 & 198.950\\
ST-Cut 3D bpj & 0.924 & 0.864 & 0.815 & 0.663 & 0.708 & 0.642 & 0.626 & 0.707 & 0.662 & 1.799 & 35.796 & 527.993\\
ST-Cut Epipolar & 0.844 & 0.772 & 0.757 & 0.759 & 0.757 & 0.753 & 0.940 & 0.740 & 0.803 & 0.013 & 1.218 & 148.860\\
Bdl.Adj.w/Sft.Mat. &0.854 & 0.782 & 0.769 & 0.769 & 0.768 & 0.766 & 0.963 & 0.760 & 0.827 & 0.011 & 0.870 & 986.089\\
CCA & 0.574 & 0.542 & 0.547 & 0.521 & 0.531 & 0.521 & 0.810 & 0.547 & 0.572 & 0.000 & 0.000 & 609.857 \\
Spectral Clustering & 0.411 & 0.275 & 0.265 & 0.272 & 0.266 & 0.264 & 0.870 & 0.257 & 0.383 & 0.007 & 0.548 & 1812.623\\
Permutation Sync. & 0.841 & 0.733 & 0.392 & 0.424 & 0.383 & 0.317 & 0.053 & 0.051 & 0.052 & 1.475 & 209.246 & 69.874\\
Fact. Graph Mat. & 0.635 & 0.473 & 0.999 & 0.431 & 0.563 & 0.431 & 0.473 & 1.000 & 0.635 & 0.000 & 0.000 & 1122.561\\
Ours w/o IQR & 0.845 & 0.745 & 0.879 & 0.579 & 0.609 & 0.567 & 0.733 & 0.571 & 0.636 & 0.787 & 30.816 & 539.157\\
Ours &  0.950&  0.908&  0.903&  0.905&  0.904&  0.902& 0.986 & 0.896 & 0.937 &0.030 & 2.511 & 371.608\\
\bottomrule
\end{tabular}
\vspace{2pt}
\centering
\caption{Quantitative comparison of all methods under Gaussian noise with standard deviation $\sigma = 1.00$.}
\label{tab:compare_noise100}
\begin{tabular}{lcccccccccccc}
\toprule
\textbf{Method} & \textbf{$\mathrm{G}$-$F_1$} & \textbf{ $\mathrm{G}$-$\mathrm{IoU}$} & \textbf{ $\mathrm{mP}$-$P$} & \textbf{ $\mathrm{mP}$-$R$} & \textbf{$\mathrm{mP}$-$F_1$} & \textbf{ $\mathrm{mP}$-$\mathrm{IoU}$} & \textbf{$\mathrm{PG}$-$P$} & \textbf{$\mathrm{PG}$-$R$}& \textbf{$\mathrm{PG}$-$F_1$}  &\textbf{$3DErr$} & \textbf{$bpe$} & {time(ms)}\\
\midrule
Greedy & 0.212 & 0.155 & 0.131 & 0.155 & 0.138 & 0.131 & 0.324 & 0.104 & 0.130 & 0.468 & 88.245 & 102.739\\
CAO & 0.603 & 0.493 & 0.382 & 0.169 & 0.234 & 0.169 & 0.559 & 0.370 & 0.382 & 0.152 & 15.164 & 196.065\\
ST-Cut 3D bpj & 0.918 & 0.853 & 0.812 & 0.651 & 0.698 & 0.630 & 0.606 & 0.691 & 0.644 & 1.715 & 38.573 & 575.972\\
ST-Cut Epipolar & 0.788 & 0.715 & 0.689 & 0.692 & 0.689 & 0.683 & 0.853 & 0.649 & 0.703 & 0.024 & 3.131 & 223.230 \\
Bdl.Adj.w/Sft.Mat. & 0.808 & 0.735 & 0.719 & 0.718 & 0.718 & 0.715 & 0.951 & 0.708 & 0.775 & 0.024 & 2.591 & 1121.568\\
CCA & 0.360 & 0.288 & 0.289 & 0.268 & 0.276 & 0.268 & 0.778 & 0.286 & 0.354 & 0.018 & 2.140 & 635.351\\
Spectral Clustering & 0.403 & 0.269 & 0.257 & 0.264 & 0.259 & 0.256 & 0.865 & 0.248 & 0.372 & 0.010 & 2.006 & 1826.594\\
Permutation Sync. & 0.841 & 0.733 & 0.393 & 0.423 & 0.382 & 0.317 & 0.053 & 0.050 & 0.051 & 1.591 & 217.600 & 65.002\\
Fact. Graph Mat. & 0.687 & 0.531 & 0.922 & 0.374 & 0.500 & 0.369 & 0.456 & 0.843 & 0.583 & 0.323 & 14.521 & 1144.011\\
Ours w/o IQR & 0.919 & 0.856 & 0.815 & 0.836 & 0.821 & 0.809 & 0.889 & 0.769 & 0.822 & 0.161 & 12.012 & 309.959\\
Ours  &0.919 & 0.856 & 0.841 & 0.840 & 0.839 & 0.835 & 0.955 & 0.821 & 0.880 & 0.087 & 14.810 & 445.098\\
\bottomrule
\end{tabular}
\vspace{2pt}
\centering
\caption{Quantitative comparison of all methods under Gaussian noise with standard deviation $\sigma = 3.00$.}
\label{tab:compare_noise300}
\begin{tabular}{lcccccccccccc}
\toprule
\textbf{Method} & \textbf{$\mathrm{G}$-$F_1$} & \textbf{ $\mathrm{G}$-$\mathrm{IoU}$} & \textbf{ $\mathrm{mP}$-$P$} & \textbf{ $\mathrm{mP}$-$R$} & \textbf{$\mathrm{mP}$-$F_1$} & \textbf{ $\mathrm{mP}$-$\mathrm{IoU}$} & \textbf{$\mathrm{PG}$-$P$} & \textbf{$\mathrm{PG}$-$R$}& \textbf{$\mathrm{PG}$-$F_1$}  &\textbf{$3DErr$} & \textbf{$bpe$} & {time(ms)}\\
\midrule
Greedy &0.211 & 0.154 & 0.131 & 0.154 & 0.138 & 0.131 & 0.325 & 0.103 & 0.129 & 0.467 & 89.223 & 103.430\\
CAO & 0.606 & 0.494 & 0.373 & 0.173 & 0.238 & 0.172 & 0.535 & 0.363 & 0.373 & 0.426 & 32.985 & 197.913\\
ST-Cut 3D bpj & 0.905 & 0.833 & 0.801 & 0.626 & 0.676 & 0.604 & 0.565 & 0.654 & 0.604 & 1.495 & 43.839 & 551.621\\
ST-Cut Epipolar & 0.678 & 0.618 & 0.596 & 0.583 & 0.586 & 0.576 & 0.751 & 0.553 & 0.585 & 0.060 & 6.809 & 328.626\\
Bdl.Adj.w/Sft.Mat. & 0.729 & 0.660 & 0.643 & 0.639 & 0.639 & 0.635 & 0.890 & 0.623 & 0.675 & 0.057 & 6.413 & 1311.430\\
CCA & 0.239 & 0.182 & 0.180 & 0.160 & 0.167 & 0.160 & 0.576 & 0.176 & 0.230 & 0.085 & 5.096 & 641.257\\
Spectral Clustering & 0.385 & 0.254 &0.244 & 0.248 & 0.244 & 0.240 & 0.854 & 0.233 & 0.352 & 0.034 & 4.592 & 1824.498\\
Permutation Sync. & 0.841 & 0.733 & 0.391 & 0.423 & 0.381 & 0.316 & 0.054 & 0.051 & 0.052 & 1.551 & 223.036 & 65.643\\
Fact. Graph Mat. &0.771 & 0.644 & 0.781 & 0.285 & 0.396 & 0.277 & 0.419 & 0.594 & 0.480 & 1.117 & 43.025 & 1123.199\\
Ours w/o IQR & 0.881 & 0.796 & 0.727 & 0.713 & 0.711 & 0.684 & 0.785 & 0.643 & 0.703 & 0.513 & 24.178 & 487.891 \\
Ours& 0.881 & 0.796 & 0.751 & 0.715 & 0.727 & 0.706 & 0.854 & 0.694 & 0.761 & 0.702 & 15.585 & 523.373\\
\bottomrule
\end{tabular}
\vspace{2pt}

\centering
\caption{Quantitative comparison of all methods under Gaussian noise with standard deviation $\sigma = 5.00$.}
\label{tab:compare_noise500}
\begin{tabular}{lcccccccccccc}
\toprule
\textbf{Method} & \textbf{$\mathrm{G}$-$F_1$} & \textbf{ $\mathrm{G}$-$\mathrm{IoU}$} & \textbf{ $\mathrm{mP}$-$P$} & \textbf{ $\mathrm{mP}$-$R$} & \textbf{$\mathrm{mP}$-$F_1$} & \textbf{ $\mathrm{mP}$-$\mathrm{IoU}$} & \textbf{$\mathrm{PG}$-$P$} & \textbf{$\mathrm{PG}$-$R$}& \textbf{$\mathrm{PG}$-$F_1$}  &\textbf{$3DErr$} & \textbf{$bpe$} & {time(ms)}\\
\midrule
Greedy & 0.200 & 0.146 & 0.123 & 0.146 & 0.130 & 0.123 & 0.298 & 0.097 & 0.119 & 0.488 & 92.961 & 104.400\\
CAO & 0.606 & 0.496 & 0.353 & 0.175 & 0.240 & 0.173 & 0.490 & 0.355 & 0.353 & 0.806 & 50.091 & 188.021 \\
ST-Cut 3D bpj & 0.897 & 0.820 & 0.797 & 0.615 & 0.667 & 0.592 & 0.533 & 0.618 & 0.570 & 1.314 & 49.919 & 601.483\\
ST-Cut Epipolar & 0.597 & 0.534 & 0.525 & 0.441 & 0.467 & 0.434 & 0.645 & 0.477 & 0.483 & 1.717 & 17.239 & 379.713\\
Bdl.Adj.w/Sft.Mat. & 0.657 & 0.582 & 0.567 & 0.550 & 0.553 & 0.543 & 0.735 & 0.523 & 0.543 & 0.463 & 16.317 & 1470.134\\
CCA & 0.215 & 0.167 & 0.161 & 0.132 & 0.143 & 0.132 & 0.470 & 0.157 & 0.202 & 0.071 & 8.209 & 636.594\\
Spectral Clustering & 0.338 & 0.219 & 0.209 & 0.205 & 0.205 & 0.199 & 0.792 & 0.191 & 0.292 & 0.109 & 10.592 & 1817.438\\
Permutation Sync. & 0.841 & 0.733 & 0.391 & 0.423 & 0.380 & 0.315 & 0.050 & 0.048 & 0.049 & 1.470 & 218.785 & 65.964\\
Fact. Graph Mat. & 0.800 & 0.689 & 0.725 & 0.264 & 0.366 & 0.255 & 0.392 & 0.487 & 0.421 & 1.373 & 78.721 & 1120.728\\
Ours w/o IQR & 0.845 & 0.745 &0.640 & 0.576 & 0.590 & 0.543 & 0.651 & 0.512 & 0.568 & 0.510 & 36.212 & 551.228 \\
Ours  &0.845 & 0.745 & 0.672 & 0.579 & 0.609 & 0.567 & 0.733 & 0.571 & 0.636 & 0.787 & 30.816 & 539.157\\
\bottomrule
\end{tabular}
\end{table*}

In this section, we conduct an ablation study to investigate the relationship between performance and the number of available camera views. We repeat the experiment using the $\sigma = 3.00$ noise benchmark, varying the number of views from 2 to 10 in increments of 1.

Interestingly, the performance initially drops as the number of camera views increases, but then begins to improve after reaching a certain point. Specifically, when increasing from 2 to 5 views, performance declines due to the inclusion of more noisy observations, which makes it harder for the method to correctly distinguish between groups. However, once the number of views exceeds 5, performance begins to recover and gradually improves up to 10 views.

This behavior can be explained by the trade-off between increased observation ambiguity and mutual consistency. As more cameras are added, the number of pairwise comparisons grows, which introduces more potential for error under noisy conditions. However, a higher number of views also brings more mutual affinity constraints, effectively acting as a voting mechanism that improves robustness to outliers.

At the 10-camera setting, the method achieves the best performance in terms of group association, with a $5.4\%$ improvement in $\mathrm{G}$-$F_1$ compared to 2-view setups. In terms of runtime, the computational cost increases polynomially with the number of views, since more point pairs must be evaluated for mutual affinity.

Moreover, reducing the number of views not only limits cross-view observations but also decreases the number of points visible in the system. While the 2-camera setup may show slightly higher accuracy for the small set of observable points, it suffers from limited perception coverage. In contrast, the 10-camera setup offers a wider field of view and higher coverage, enabling better overall association despite the increased complexity.

\subsection{Comparison With Existing Works}

In this experiment, we compared our method against feature-free spatial geometry baselines, which rely solely on 2D point observations and camera poses as input. While some of the comparison methods are relatively dated, this is due to the underexplored nature of the problem addressed in this paper. These classical approaches are rarely updated, but they continue to serve as foundational modules for solving more complex tasks.

We selected methods that operate purely on spatial geometry, as discussed in Section~\ref{sec:related_work_trainingfree}. These methods take the camera poses and 2D point coordinates in the image plane as input and produce association results in the form of grouped 2D points. The compared baselines include: the Greedy method~\cite{shafique2005noniterative}, Composition-based Affinity Optimization (CAO)~\cite{swoboda2019convex}, ST-cut using 3D back-projection constraints (ST-Cut 3D bpj)~\cite{vogiatzis2005multi}, ST-cut using epipolar constraints (ST-Cut Epipolar)~\cite{kolmogorov2002multi}, bundle adjustment with soft matching (Bdl.Adj. w/ Sft. Mat.)~\cite{zach2014robust}, Connected Component Analysis (CCA)~\cite{klasing2008clustering}, Spectral Clustering~\cite{li2015large}, Permutation Synchronization (Permutation Sync.)~\cite{li2022fast}, and Factorized Graph Matching (Fact. Graph Mat.)~\cite{zhou2015factorized}. We also reported results for our proposed method under three different configurations: the original full model, and a variant without IQR-based outlier removal.

We conducted experiments under four different levels of Gaussian noise: $\sigma=0.00$, $\sigma=1.00$, $\sigma=3.00$, and $\sigma=5.00$. The results are summarized in Tables~\ref{tab:compare_noise000}, \ref{tab:compare_noise100}, \ref{tab:compare_noise300}, and \ref{tab:compare_noise500}. For each batch (i.e., scene), evaluation metrics were computed individually and then averaged across all batches.

Under noise-free conditions ($\sigma=0.00$), our method achieves a $13.3\%$ improvement over the second-best method in perfect group association ($PG$-$F_1$), a $2.8\%$ improvement in imperfect group association ($G$-$F_1$), and a $17.7\%$ gain in point-wise association performance ($mP$-$F_1$). In terms of $\mathrm{IoU}$, our method outperforms the second-best by $5.1\%$ for imperfect groups ($G$-$\mathrm{IoU}$) and $17.5\%$ for point-wise association quality ($mp$-$\mathrm{IoU}$). Under $\sigma=1.00$ conditions, our method achieves a $10.5\%$ improvement over the second-best method in $PG$-$F_1$, a $16.6\%$ gain in $mP$-$F_1$, a $0.1\%$ gain in $mP$-$F_1$, an $0.3\%$ gain in ($G$-$\mathrm{IoU}$), and an $16.8\%$ gain in ($mP$-$\mathrm{IoU}$). Under $\sigma=3.00$ conditions, our method achieves a $12.7\%$ improvement over the second-best method in $PG$-$F_1$, a $7.5\%$ gain in $mP$-$F_1$, and an $11.1\%$ gain in ($mP$-$\mathrm{IoU}$).

While our method ranks mid-range in 3D reconstruction error and BPE, this is largely due to the conservative nature of several baselines. These methods tend to produce fewer or no associations under point-rich conditions, leading to a higher likelihood of skipping difficult cases. As a result, they often report lower mean 3D error and BPE, since missing group associations do not incur penalties in the error calculation.

\begin{figure*}
    \centering
   \centerline{\includegraphics[width=\linewidth]{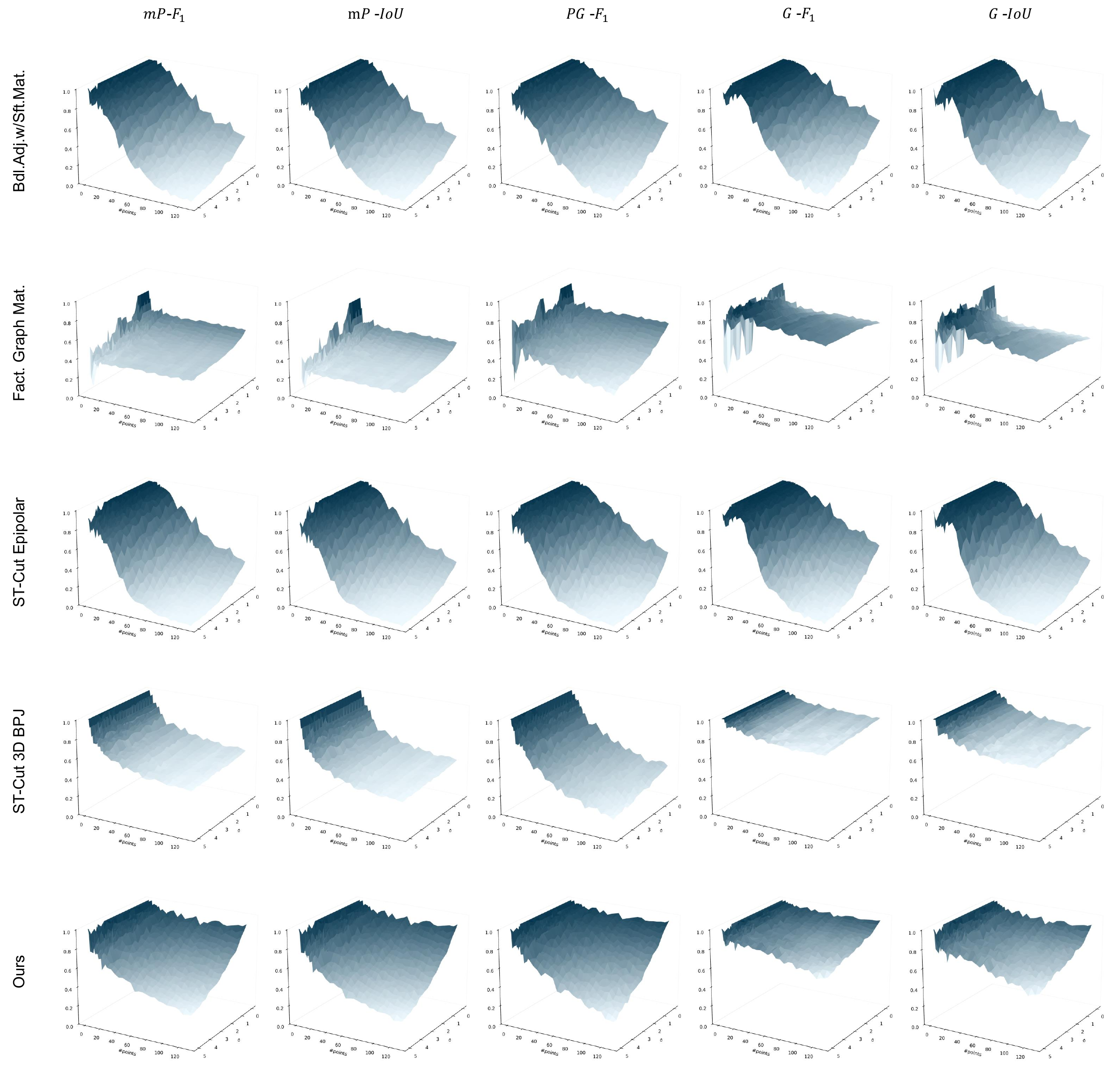}}
    \caption{Accuracy comparison of various baselines across varying noise levels and object instance counts.}
    \label{fig:map3d}
\end{figure*}

We provided qualitative comparisons using 3D visualizations to compare the top four performing baseline methods with our proposed approach under varying numbers of points (\# points) and different noise levels ($\sigma$). Ideally, a well-performing method should produce 3D score maps with a flat, elevated surface at $z=1$, indicating consistent and accurate associations. These visual results are summarized in Fig.~\ref{fig:map3d}. Our method consistently outperformed the baselines in both point-wise and group associations. The $mP$ score maps generated by our method were notably flatter and higher than those of the baselines, especially under low noise conditions. Moreover, our method exhibited more gradual performance degradation as the number of points and noise levels increased. In terms of perfect group association ($\mathrm{PG}$-$F_1$), our method maintained relatively high and stable performance, with a slower decline across increasing noise and point density. The imperfect group association scores ($\mathrm{G}$-$F_1$) also demonstrated a flat plateau near the top surface, further highlighting the robustness of our method. These results collectively indicated that our approach offered superior robustness and accuracy compared to baseline methods under varying noise and scene complexity.

\begin{figure*}
    \centering
   \centerline{\includegraphics[width=\linewidth]{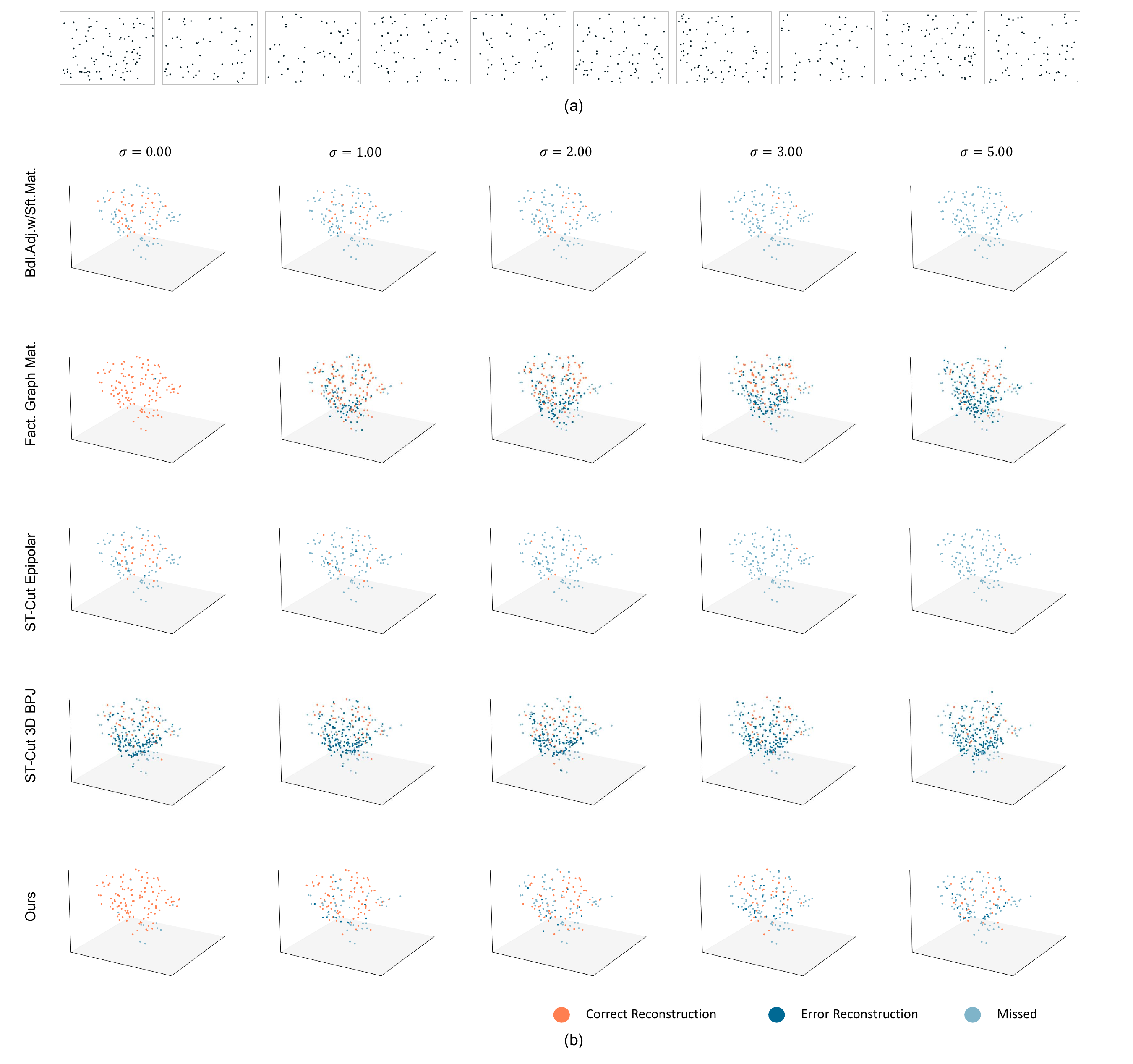}}
    \caption{Comparison of 3D reconstruction results for 130 instances across various baselines under different noise levels. (a) 10 observision views with visualized 2D points ($\sigma=0.00$). (b) 3D reconstruction results.}
    \label{fig:recresult}
\end{figure*}

We used the association results from various baseline methods and our proposed approach to perform 3D reconstruction, and we summarized the visualization results in Fig.~\ref{fig:recresult}. A scene containing 130 instances was randomly selected from the dataset for this experiment. This visualization provides an intuitive understanding of how many points are correctly reconstructed, how many are incorrectly reconstructed, and how many are missed entirely. An ideal method would accurately reconstruct all points without introducing errors or omissions. As shown in the figure, our method performs well under noise-free conditions, missing only a few points. Compared to the baselines, our approach yields more accurate association groups with fewer incorrect associations. Notably, the number of missed associations is significantly higher than that of incorrect ones, which is desirable in challenging, high-noise conditions where avoiding false positives is more critical than recovering every point. This characteristic is beneficial for downstream tasks, as it reduces the introduction of noisy data into subsequent processing stages.

\begin{figure}
    \centering
   \centerline{\includegraphics[width=\linewidth]{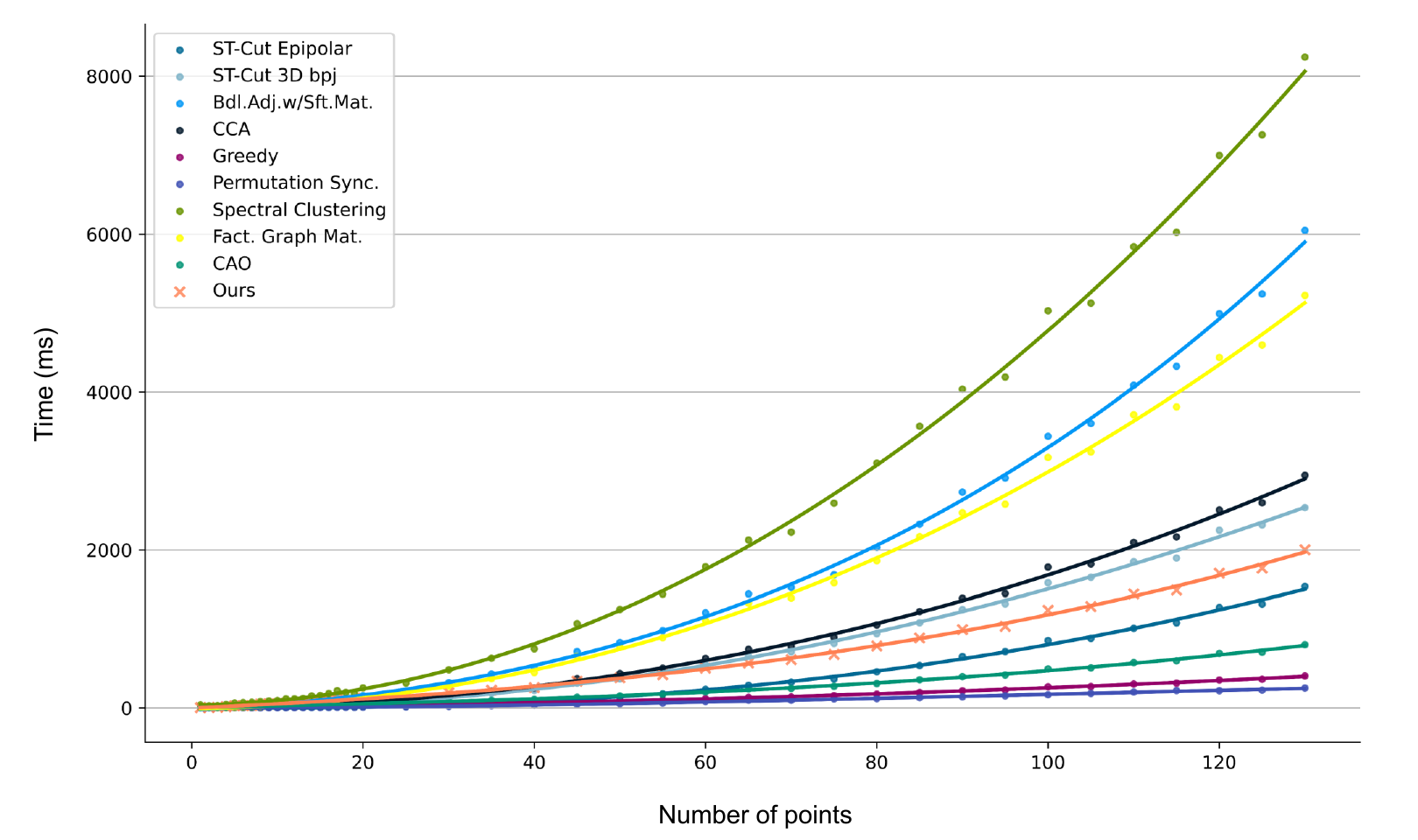}}
    \caption{Comparison of time cost across different methods under varying object instance counts.}
    \label{fig:time}
\end{figure}

Finally, we compare the runtime performance of our method against the baselines, as shown in Fig.~\ref{fig:time}. The time cost for each method is computed by averaging the runtime over all batches. As the number of points increases, the computational cost of most baseline methods rises rapidly, often exhibiting polynomial growth. In particular, methods that rely on pairwise affinity scores tend to show approximately quadratic scaling in both time and memory complexity. Although some local-optimum-based methods are more computationally efficient, their performance—as discussed in previous sections—is significantly worse in terms of accuracy. In contrast, our method demonstrates favorable scalability, maintaining a competitive runtime while achieving superior association performance.

\section{Discussion}

Our method demonstrates consistent improvements over geometry-based baselines across various noise levels and numbers of point instances. In addition to its accuracy, our method shows favorable runtime performance, achieving a strong balance between efficiency and accuracy. It performs particularly well in noisy and instance-rich environments by effectively leveraging mutual affinities, leading to cleaner association results and robust correspondence under challenging conditions. In noise-free or low-noise scenarios, the method remains reliable and achieves high scores in the dense-point setups.

We observed that the method initially suffers when the number of camera views increases under high-noise conditions, due to greater ambiguity introduced by additional noisy observations. However, as the number of views increases beyond a certain threshold, performance improves again, benefiting from more consistent affinity information. This behavior highlights the importance of mutual voting in improving robustness against noise.

A key strength of our method is its robustness without relying on visual features, making it particularly well-suited for texture-poor or featureless scenarios. In addition, this makes it ideal as an intermediate module that bridges upstream and downstream tasks—for example, serving as a correspondence associator between multi-view object detection and 3D reconstruction.

This method was originally developed for a real-time 3D visualization system in laparoscopic surgery training~\cite{sun2024easyvis2}, where a dense 10-camera setup is used to observe object interactions. In that context, our method functions as a multi-view, multi-object associator between detection outputs and downstream 3D reconstruction, even in the absence of distinctive object features. The camera poses in this system are calibrated using SfM, which motivated our design choice to leverage known poses and operate solely on 2D point observations.

While the method demonstrates high accuracy and runtime efficiency in this configuration, its time cost increases polynomially with the number of object instances. Future work will focus on enhancing scalability for high-instance-count scenes and improving runtime performance for real-time deployment.

\section{Conclusion}
We proposed an efficient method, C-DOG, for multi-view, multi-object association using only 2D point observations and known camera poses. Designed to operate effectively in featureless and noisy environments, our approach eliminates the need for visual features while preserving both scalability and accuracy. Extensive evaluations across varying noise levels, object densities, and view configurations demonstrate that our method consistently outperforms geometry-based baselines, achieving a strong balance between precision and runtime. This provides a general and robust solution that can serve as a bridging module between multi-view detection and 3D reconstruction. Future work will focus on improving computational efficiency and extending the approach to real-time applications in instance-rich environments.

\section*{Acknowledgment}
This work was supported by the National Institute of Biomedical Imaging and Bioengineering (NIBIB) of the U.S. National Institutes of Health (NIH) under award number R01EB019460.

\bibliographystyle{ieeetr} 
\bibliography{reference}

\begin{thebibliography}{10}

\bibitem{furukawa2015multi}
Y.~Furukawa, C.~Hern{\'a}ndez, {\em et~al.}, ``Multi-view stereo: A tutorial,'' {\em Foundations and trends{\textregistered} in Computer Graphics and Vision}, vol.~9, no.~1-2, pp.~1--148, 2015.

\bibitem{kubota2007multiview}
A.~Kubota, A.~Smolic, M.~Magnor, M.~Tanimoto, T.~Chen, and C.~Zhang, ``Multiview imaging and 3dtv,'' {\em IEEE signal processing magazine}, vol.~24, no.~6, pp.~10--21, 2007.

\bibitem{teepe2024lifting}
T.~Teepe, P.~Wolters, J.~Gilg, F.~Herzog, and G.~Rigoll, ``Lifting multi-view detection and tracking to the bird's eye view,'' in {\em Proceedings of the IEEE/CVF Conference on Computer Vision and Pattern Recognition}, pp.~667--676, 2024.

\bibitem{kunert20133d}
W.~Kunert, P.~Storz, and A.~Kirschniak, ``For 3d laparoscopy: a step toward advanced surgical navigation: how to get maximum benefit from 3d vision,'' {\em Surgical endoscopy}, vol.~27, no.~2, pp.~696--699, 2013.

\bibitem{aulinas2008slam}
J.~Aulinas, Y.~Petillot, J.~Salvi, and X.~Llad{\'o}, ``The slam problem: a survey,'' {\em Artificial Intelligence Research and Development}, pp.~363--371, 2008.

\bibitem{huang2024survey}
Q.~Huang, X.~Guo, Y.~Wang, H.~Sun, and L.~Yang, ``A survey of feature matching methods,'' {\em IET Image Processing}, vol.~18, no.~6, pp.~1385--1410, 2024.

\bibitem{szeliski2022computer}
R.~Szeliski, {\em Computer vision: algorithms and applications}.
\newblock Springer Nature, 2022.

\bibitem{ji2024view}
D.~Ji, S.~Gao, L.~Zhu, Q.~Zhu, Y.~Zhao, P.~Xu, H.~Lu, F.~Zhao, and J.~Ye, ``View-centric multi-object tracking with homographic matching in moving uav,'' {\em arXiv preprint arXiv:2403.10830}, 2024.

\bibitem{zhang2025collaborative}
Z.~Zhang, F.~Shi, C.~Jia, M.~Zhao, and X.~Cheng, ``Collaborative association network for multi-view multi-human association and tracking using constraint optimization and object search,'' in {\em ICASSP 2025-2025 IEEE International Conference on Acoustics, Speech and Signal Processing (ICASSP)}, pp.~1--5, IEEE, 2025.

\bibitem{shuai2022adaptive}
H.~Shuai, L.~Wu, and Q.~Liu, ``Adaptive multi-view and temporal fusing transformer for 3d human pose estimation,'' {\em IEEE Transactions on Pattern Analysis and Machine Intelligence}, vol.~45, no.~4, pp.~4122--4135, 2022.

\bibitem{zhang2021direct}
J.~Zhang, Y.~Cai, S.~Yan, J.~Feng, {\em et~al.}, ``Direct multi-view multi-person 3d pose estimation,'' {\em Advances in Neural Information Processing Systems}, vol.~34, pp.~13153--13164, 2021.

\bibitem{zhou2023efficient}
K.~Zhou, L.~Zhang, F.~Lu, X.-D. Zhou, and Y.~Shi, ``Efficient hierarchical multi-view fusion transformer for 3d human pose estimation,'' in {\em Proceedings of the 31st ACM International Conference on Multimedia}, pp.~7512--7520, 2023.

\bibitem{ma2022ppt}
H.~Ma, Z.~Wang, Y.~Chen, D.~Kong, L.~Chen, X.~Liu, X.~Yan, H.~Tang, and X.~Xie, ``Ppt: token-pruned pose transformer for monocular and multi-view human pose estimation,'' in {\em European Conference on Computer Vision}, pp.~424--442, Springer, 2022.

\bibitem{zhang2021adafuse}
Z.~Zhang, C.~Wang, W.~Qiu, W.~Qin, and W.~Zeng, ``Adafuse: Adaptive multiview fusion for accurate human pose estimation in the wild,'' {\em International Journal of Computer Vision}, vol.~129, pp.~703--718, 2021.

\bibitem{yu2022towards}
E.~Yu, Z.~Li, and S.~Han, ``Towards discriminative representation: Multi-view trajectory contrastive learning for online multi-object tracking,'' in {\em Proceedings of the IEEE/CVF conference on computer vision and pattern recognition}, pp.~8834--8843, 2022.

\bibitem{dong2021shape}
Z.~Dong, J.~Song, X.~Chen, C.~Guo, and O.~Hilliges, ``Shape-aware multi-person pose estimation from multi-view images,'' in {\em Proceedings of the IEEE/CVF International Conference on Computer Vision}, pp.~11158--11168, 2021.

\bibitem{cai2020messytable}
Z.~Cai, J.~Zhang, D.~Ren, C.~Yu, H.~Zhao, S.~Yi, C.~K. Yeo, and C.~Change~Loy, ``Messytable: Instance association in multiple camera views,'' in {\em Computer Vision--ECCV 2020: 16th European Conference, Glasgow, UK, August 23--28, 2020, Proceedings, Part XI 16}, pp.~1--16, Springer, 2020.

\bibitem{dong2019fast}
J.~Dong, W.~Jiang, Q.~Huang, H.~Bao, and X.~Zhou, ``Fast and robust multi-person 3d pose estimation from multiple views,'' in {\em Proceedings of the IEEE/CVF conference on computer vision and pattern recognition}, pp.~7792--7801, 2019.

\bibitem{chen2014near}
Y.~Chen, L.~J. Guibas, and Q.-X. Huang, ``Near-optimal joint object matching via convex relaxation,'' {\em arXiv preprint arXiv:1402.1473}, 2014.

\bibitem{fathian2020clear}
K.~Fathian, K.~Khosoussi, Y.~Tian, P.~Lusk, and J.~P. How, ``Clear: A consistent lifting, embedding, and alignment rectification algorithm for multiview data association,'' {\em IEEE Transactions on Robotics}, vol.~36, no.~6, pp.~1686--1703, 2020.

\bibitem{li2022fast}
S.~Li, Y.~Shi, and G.~Lerman, ``Fast, accurate and memory-efficient partial permutation synchronization,'' in {\em Proceedings of the IEEE/CVF Conference on Computer Vision and Pattern Recognition}, pp.~15735--15743, 2022.

\bibitem{zhou2015factorized}
F.~Zhou and F.~De~la Torre, ``Factorized graph matching,'' {\em IEEE transactions on pattern analysis and machine intelligence}, vol.~38, no.~9, pp.~1774--1789, 2015.

\bibitem{kahl2025towards}
M.~Kahl, S.~Stricker, L.~Hutschenreiter, F.~Bernard, C.~Rother, and B.~Savchynskyy, ``Towards optimizing large-scale multi-graph matching in bioimaging,'' in {\em Proceedings of the Computer Vision and Pattern Recognition Conference}, pp.~11569--11578, 2025.

\bibitem{li2015large}
Y.~Li, F.~Nie, H.~Huang, and J.~Huang, ``Large-scale multi-view spectral clustering via bipartite graph,'' in {\em Proceedings of the AAAI conference on artificial intelligence}, vol.~29, 2015.

\bibitem{meghanathan2016greedy}
N.~Meghanathan, ``A greedy algorithm for neighborhood overlap-based community detection,'' {\em Algorithms}, vol.~9, no.~1, p.~8, 2016.

\bibitem{cetinkayaRundel2024IMS}
M.~Çetinkaya{-}Rundel and J.~Hardin, {\em Introduction to Modern Statistics}.
\newblock OpenIntro, 2~ed., 2024.
\newblock Open-access textbook.

\bibitem{rublee2011orb}
E.~Rublee, V.~Rabaud, K.~Konolige, and G.~Bradski, ``Orb: An efficient alternative to sift or surf,'' in {\em 2011 International conference on computer vision}, pp.~2564--2571, Ieee, 2011.

\bibitem{fischler1981random}
M.~A. Fischler and R.~C. Bolles, ``Random sample consensus: a paradigm for model fitting with applications to image analysis and automated cartography,'' {\em Communications of the ACM}, vol.~24, no.~6, pp.~381--395, 1981.

\bibitem{detone2018superpoint}
D.~DeTone, T.~Malisiewicz, and A.~Rabinovich, ``Superpoint: Self-supervised interest point detection and description,'' in {\em Proceedings of the IEEE conference on computer vision and pattern recognition workshops}, pp.~224--236, 2018.

\bibitem{sarlin2020superglue}
P.-E. Sarlin, D.~DeTone, T.~Malisiewicz, and A.~Rabinovich, ``Superglue: Learning feature matching with graph neural networks,'' in {\em Proceedings of the IEEE/CVF conference on computer vision and pattern recognition}, pp.~4938--4947, 2020.

\bibitem{shi2022clustergnn}
Y.~Shi, J.-X. Cai, Y.~Shavit, T.-J. Mu, W.~Feng, and K.~Zhang, ``Clustergnn: Cluster-based coarse-to-fine graph neural network for efficient feature matching,'' in {\em Proceedings of the IEEE/CVF conference on computer vision and pattern recognition}, pp.~12517--12526, 2022.

\bibitem{zhang2025comatcher}
J.~Zhang, Z.~Xia, M.~Dong, S.~Shen, L.~Yue, and X.~Zheng, ``Comatcher: Multi-view collaborative feature matching,'' in {\em Proceedings of the Computer Vision and Pattern Recognition Conference}, pp.~21970--21980, 2025.

\bibitem{roessle2023end2end}
B.~Roessle and M.~Nie{\ss}ner, ``End2end multi-view feature matching with differentiable pose optimization,'' in {\em Proceedings of the IEEE/CVF International Conference on Computer Vision}, pp.~477--487, 2023.

\bibitem{maset2017practical}
E.~Maset, F.~Arrigoni, and A.~Fusiello, ``Practical and efficient multi-view matching,'' in {\em Proceedings of the IEEE International Conference on Computer Vision}, pp.~4568--4576, 2017.

\bibitem{wen2023unpaired}
Y.~Wen, S.~Wang, Q.~Liao, W.~Liang, K.~Liang, X.~Wan, and X.~Liu, ``Unpaired multi-view graph clustering with cross-view structure matching,'' {\em IEEE Transactions on Neural Networks and Learning Systems}, 2023.

\bibitem{vaswani2017attention}
A.~Vaswani, N.~Shazeer, N.~Parmar, J.~Uszkoreit, L.~Jones, A.~N. Gomez, {\L}.~Kaiser, and I.~Polosukhin, ``Attention is all you need,'' {\em Advances in neural information processing systems}, vol.~30, 2017.

\bibitem{dosovitskiy2020image}
A.~Dosovitskiy, L.~Beyer, A.~Kolesnikov, D.~Weissenborn, X.~Zhai, T.~Unterthiner, M.~Dehghani, M.~Minderer, G.~Heigold, S.~Gelly, {\em et~al.}, ``An image is worth 16x16 words: Transformers for image recognition at scale,'' {\em arXiv preprint arXiv:2010.11929}, 2020.

\bibitem{scarselli2008graph}
F.~Scarselli, M.~Gori, A.~C. Tsoi, M.~Hagenbuchner, and G.~Monfardini, ``The graph neural network model,'' {\em IEEE transactions on neural networks}, vol.~20, no.~1, pp.~61--80, 2008.

\bibitem{wu2020comprehensive}
Z.~Wu, S.~Pan, F.~Chen, G.~Long, C.~Zhang, and P.~S. Yu, ``A comprehensive survey on graph neural networks,'' {\em IEEE transactions on neural networks and learning systems}, vol.~32, no.~1, pp.~4--24, 2020.

\bibitem{wu2021graph}
S.~Wu, S.~Jin, W.~Liu, L.~Bai, C.~Qian, D.~Liu, and W.~Ouyang, ``Graph-based 3d multi-person pose estimation using multi-view images,'' in {\em Proceedings of the IEEE/CVF international conference on computer vision}, pp.~11148--11157, 2021.

\bibitem{luna2022graph}
E.~Luna, J.~C. SanMiguel, J.~M. Mart{\'\i}nez, and P.~Carballeira, ``Graph neural networks for cross-camera data association,'' {\em IEEE Transactions on Circuits and Systems for Video Technology}, vol.~33, no.~2, pp.~589--601, 2022.

\bibitem{rodriguez2024multi}
D.~Rodriguez-Criado, P.~Bachiller-Burgos, G.~Vogiatzis, and L.~J. Manso, ``Multi-person 3d pose estimation from unlabelled data,'' {\em Machine Vision and Applications}, vol.~35, no.~3, p.~46, 2024.

\bibitem{shafique2005noniterative}
K.~Shafique and M.~Shah, ``A noniterative greedy algorithm for multiframe point correspondence,'' {\em IEEE transactions on pattern analysis and machine intelligence}, vol.~27, no.~1, pp.~51--65, 2005.

\bibitem{lu2004wide}
X.~Lu and R.~Manduchi, ``Wide baseline feature matching using the cross-epipolar ordering constraint,'' in {\em Proceedings of the 2004 IEEE Computer Society Conference on Computer Vision and Pattern Recognition, 2004. CVPR 2004.}, vol.~1, pp.~I--I, IEEE, 2004.

\bibitem{kolmogorov2002multi}
V.~Kolmogorov and R.~Zabih, ``Multi-camera scene reconstruction via graph cuts,'' in {\em Computer Vision—ECCV 2002: 7th European Conference on Computer Vision Copenhagen, Denmark, May 28--31, 2002 Proceedings, Part III 7}, pp.~82--96, Springer, 2002.

\bibitem{vogiatzis2005multi}
G.~Vogiatzis, P.~H. Torr, and R.~Cipolla, ``Multi-view stereo via volumetric graph-cuts,'' in {\em 2005 IEEE Computer Society Conference on Computer Vision and Pattern Recognition (CVPR'05)}, vol.~2, pp.~391--398, IEEE, 2005.

\bibitem{pachauri2013solving}
D.~Pachauri, R.~Kondor, and V.~Singh, ``Solving the multi-way matching problem by permutation synchronization,'' {\em Advances in neural information processing systems}, vol.~26, 2013.

\bibitem{yan2015multi}
J.~Yan, M.~Cho, H.~Zha, X.~Yang, and S.~M. Chu, ``Multi-graph matching via affinity optimization with graduated consistency regularization,'' {\em IEEE transactions on pattern analysis and machine intelligence}, vol.~38, no.~6, pp.~1228--1242, 2015.

\bibitem{swoboda2019convex}
P.~Swoboda, A.~Mokarian, C.~Theobalt, F.~Bernard, {\em et~al.}, ``A convex relaxation for multi-graph matching,'' in {\em Proceedings of the IEEE/CVF Conference on Computer Vision and Pattern Recognition}, pp.~11156--11165, 2019.

\bibitem{zach2014robust}
C.~Zach, ``Robust bundle adjustment revisited,'' in {\em Computer Vision--ECCV 2014: 13th European Conference, Zurich, Switzerland, September 6-12, 2014, Proceedings, Part V 13}, pp.~772--787, Springer, 2014.

\bibitem{zhang2024rkhs}
R.~Zhang, J.~Song, X.~Gao, J.~Wu, T.~Liu, J.~Zhang, R.~Eustice, and M.~Ghaffari, ``Rkhs-ba: A robust correspondence-free multi-view registration framework with semantic point clouds,'' {\em arXiv preprint arXiv:2403.01254}, 2024.

\bibitem{ng2001spectral}
A.~Ng, M.~Jordan, and Y.~Weiss, ``On spectral clustering: Analysis and an algorithm,'' {\em Advances in neural information processing systems}, vol.~14, 2001.

\bibitem{klasing2008clustering}
K.~Klasing, D.~Wollherr, and M.~Buss, ``A clustering method for efficient segmentation of 3d laser data,'' in {\em 2008 IEEE international conference on robotics and automation}, pp.~4043--4048, IEEE, 2008.

\bibitem{cao2017realtime}
Z.~Cao, T.~Simon, S.-E. Wei, and Y.~Sheikh, ``Realtime multi-person 2d pose estimation using part affinity fields,'' in {\em Proceedings of the IEEE conference on computer vision and pattern recognition}, pp.~7291--7299, 2017.

\bibitem{hennig2015handbook}
C.~Hennig, M.~Meila, F.~Murtagh, and R.~Rocci, {\em Handbook of cluster analysis}.
\newblock CRC press, 2015.

\bibitem{belagiannis20143d}
V.~Belagiannis, S.~Amin, M.~Andriluka, B.~Schiele, N.~Navab, and S.~Ilic, ``3d pictorial structures for multiple human pose estimation,'' in {\em Proceedings of the IEEE conference on computer vision and pattern recognition}, pp.~1669--1676, 2014.

\bibitem{joo2015panoptic}
H.~Joo, H.~Liu, L.~Tan, L.~Gui, B.~Nabbe, I.~Matthews, T.~Kanade, S.~Nobuhara, and Y.~Sheikh, ``Panoptic studio: A massively multiview system for social motion capture,'' in {\em Proceedings of the IEEE international conference on computer vision}, pp.~3334--3342, 2015.

\bibitem{sun2024easyvis2}
Y.-H. Sun, G.~Shen, J.~Chen, J.~Fernandes, H.~Jiang, and Y.~H. Hu, ``Easyvis2: A real time multi-view 3d visualization for laparoscopic surgery training enhanced by a deep neural network yolov8-pose,'' {\em arXiv preprint arXiv:2412.16742}, 2024.

\bibitem{sun2025easyvis}
Y.-H. Sun, J.~Ke, J.~Fernandes, J.~Chen, H.~Jiang, and Y.~H. Hu, ``Easyvis: a real-time 3d visualization software system for laparoscopic surgery box trainer,'' {\em Updates in Surgery}, pp.~1--16, 2025.

\bibitem{moulon2017openmvg}
P.~Moulon, P.~Monasse, R.~Perrot, and R.~Marlet, ``Openmvg: Open multiple view geometry,'' in {\em Reproducible Research in Pattern Recognition: First International Workshop, RRPR 2016, Canc{\'u}n, Mexico, December 4, 2016, Revised Selected Papers 1}, pp.~60--74, Springer, 2017.

\end{thebibliography}

\end{document}